\documentclass[10pt,twocolumn,letterpaper]{article}

\usepackage{cvpr}
\usepackage{times}
\usepackage{epsfig}
\usepackage{amsmath}
\usepackage{amssymb}
\usepackage{enumitem}
\usepackage[ruled]{algorithm2e}
\usepackage{array}
\usepackage{multirow}
\usepackage{color,graphicx}
\usepackage{cases}
\usepackage[export]{adjustbox}
\usepackage[dvipsnames]{xcolor}
\usepackage{textpos}
\usepackage{siunitx}

\usepackage[pagebackref=true,breaklinks=true,letterpaper=true,colorlinks,bookmarks=false]{hyperref}

\cvprfinalcopy % *** Uncomment this line for the final submission

\def\cvprPaperID{7487} % *** Enter the CVPR Paper ID here

\ifcvprfinal\fi

\DeclareMathOperator*{\argmin}{argmin}
\providecommand{\tabularnewline}{\\}

\begin{document}

%%%%%%%%% TITLE
\title{A Context-Aware Loss Function for Action Spotting in Soccer Videos}

\author{
Anthony Cioppa*\\
{\small University of Li\`ege}\\
%Universit\'e de Li\`ege\\
%Liège, Brussels\\
{\tt\small anthony.cioppa@uliege.be}
\and
Adrien Deli\`ege*\\
{\small University of Li\`ege}\\
%Universit\'e de Li\`ege\\
%Liège, Brussels\\
{\tt\small adrien.deliege@uliege.be}
\and
Silvio Giancola*\\
{\small KAUST}\\
%Thuwal, KSA\\
{\tt\small silvio.giancola@kaust.edu.sa}
\and
Bernard Ghanem\\
{\small KAUST}\\
%Thuwal, KSA\\
%{\tt\small bernard.ghanem@kaust.edu.sa}
\and
Marc Van Droogenbroeck\\
{\small University of Li\`ege}\\
%Universit\'e de Li\`ege\\
%Liège, Brussels\\
%{\tt\small M.VanDroogenbroeck@uliege.be}
\and
Rikke Gade\\
{\small Aalborg University}\\
%Danmark\\
%{\tt\small mail@mail.dk}
\and
Thomas B. Moeslund\\
{\small Aalborg University}\\
%Danmark\\
%{\tt\small mail@mail.dk}
%\and
%$^*$These authors contributed equally.
}

\maketitle

%\thispagestyle{empty}
% Custom useful commands
\newcommand{\mysection}[1]{\vspace{2pt}\noindent\textbf{#1}}
\newcommand{\Table}[1]{Table~\ref{tab:#1}}
\newcommand{\Figure}[1]{Figure~\ref{fig:#1}}
\newcommand{\Equation}[1]{Equation~\eqref{eq:#1}}
\newcommand{\Equations}[2]{Equations \eqref{eq:#1} and \eqref{eq:#2}}
\newcommand{\Section}[1]{Section~\ref{sec:#1}}
\newcommand{\SoccerNet}{SoccerNet~\cite{Giancola_2018_CVPR_Workshops}\xspace}
\newcommand{\ActivityNet}{ActivityNet~\cite{caba2015activitynet}\xspace}

\newcommand\blfootnote[1]{%
  \begingroup
  \renewcommand\thefootnote{}\footnote{#1}%
  \addtocounter{footnote}{-1}%
  \endgroup
}
\blfootnote{\textbf{(*)} Denotes equal contributions. Code available at \url{https://github.com/cioppaanthony/context-aware-loss}.}

\definecolor{myred}[a=.5]{RGB}{215,25,28} 
\definecolor{myorange}[a=.5]{RGB}{253,174,97}
\definecolor{anthoblue}[a=.5]{RGB}{31,119,180}
\definecolor{anthoorange}[a=.5]{RGB}{255,127,14}
\definecolor{anthogreen}[a=.5]{RGB}{0,150,0}
\definecolor{anthored}[a=.5]{RGB}{150,0,0}
\definecolor{anthobrown}[a=.5]{RGB}{153,76,0}
\definecolor{mygreen}[a=.5]{RGB}{166,217,106} 
\definecolor{mygray}[a=.5]{gray}{0.6}

\newcommand{\whitebox}{\hfill\textcolor{white}{\rule[1mm]{1.8mm}{2.8mm}}\hfill}
\newcommand{\redbox}{\hfill\textcolor{myred}{\rule[1mm]{1.8mm}{2.8mm}}\hfill}
\newcommand{\orangebox}{\hfill\textcolor{myorange}{\rule[1mm]{1.8mm}{2.8mm}}\hfill}
\newcommand{\greenbox}{\hfill\textcolor{mygreen}{\rule[1mm]{1.8mm}{2.8mm}}\hfill}
\newcommand{\graybox}{\hfill\textcolor{mygray}{\rule[1mm]{1.8mm}{2.8mm}}\hfill}

%%%%%%%%% ABSTRACT
\begin{abstract}
In video understanding, action spotting consists in temporally localizing human-induced events annotated with single timestamps. In this paper, we propose a novel loss function that specifically considers the temporal context naturally present around each action, rather than focusing on the single annotated frame to spot. We benchmark our loss on a large dataset of soccer videos, SoccerNet, and achieve an improvement of  $12.8\%$ over the baseline. We show the generalization capability of our loss for generic activity proposals and detection on ActivityNet, by spotting the beginning and the end of each activity.  Furthermore, we provide an extended ablation study and display challenging cases for action spotting in soccer videos. Finally, we qualitatively illustrate how our loss induces a precise temporal understanding of actions and show how such semantic knowledge can be used for automatic highlights generation.
    
    %% ORIGINAL
    %Action spotting is an important element of general activity understanding. It consists of detecting human-induced events annotated with single timestamps. In this paper, we propose a novel loss function for action spotting. Our loss aims at dealing specifically with the temporal context naturally present around an action. Rather than focusing on the single annotated frame of the action to spot, we consider different temporal segments surrounding it and shape our loss function accordingly. We test our loss on SoccerNet, a large dataset of soccer videos, showing an improvement of $12.8\%$ on the current baseline. We also show the generalization capability of our loss function on ActivityNet for activity proposals and detection, by spotting the beginning and the end of each activity. Furthermore, we provide an extended ablation study and identify challenging cases for action spotting in soccer videos. Finally, we qualitatively illustrate how our loss induces a precise temporal understanding of actions, and how such semantic knowledge can be leveraged to design a highlights generator.
\end{abstract}

%%%%%%%%% BODY TEXT

\section{Introduction}
\label{sec:Intro}

%  Teaser Figure
\begin{figure}[t]
    \centering
    % From https://tex.stackexchange.com/questions/167940/a-movie-film-strip-of-images
    \setlength{\fboxsep}{0pt}%
    \colorbox{black}{%
        \begin{minipage}{\linewidth}
            \rule{0mm}{4.8mm}
            \redbox\redbox\redbox\redbox\orangebox
            \orangebox\graybox\graybox\graybox\graybox
            \greenbox\greenbox\greenbox\greenbox\orangebox
            \orangebox\redbox\redbox\redbox\redbox
            \null\\
            \null\hfill
            \includegraphics[width=0.235\textwidth]{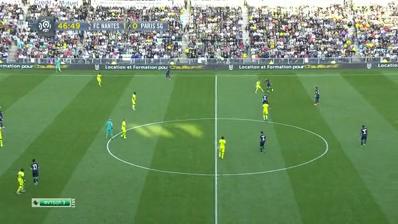}
            \includegraphics[width=0.235\textwidth]{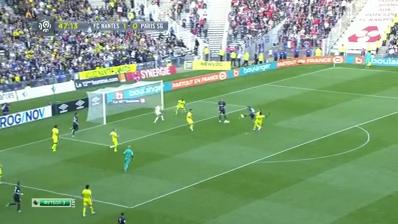}
            \includegraphics[width=0.235\textwidth]{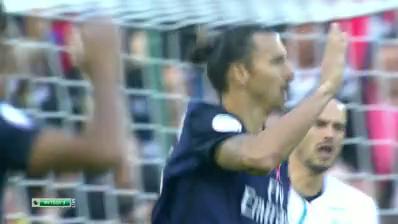}
            \includegraphics[width=0.235\textwidth]{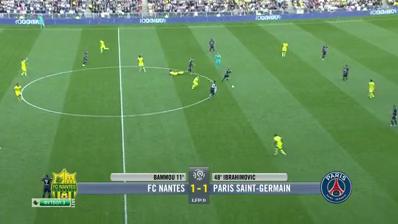}
            \null\hfill
            \\[1mm]%
            \null
            \redbox\redbox\redbox\redbox\orangebox
            \orangebox\graybox\graybox\graybox\graybox
            \greenbox\greenbox\greenbox\greenbox\orangebox
            \orangebox\redbox\redbox\redbox\redbox
            \null
        \end{minipage}
        }
    \includegraphics[width=\linewidth]{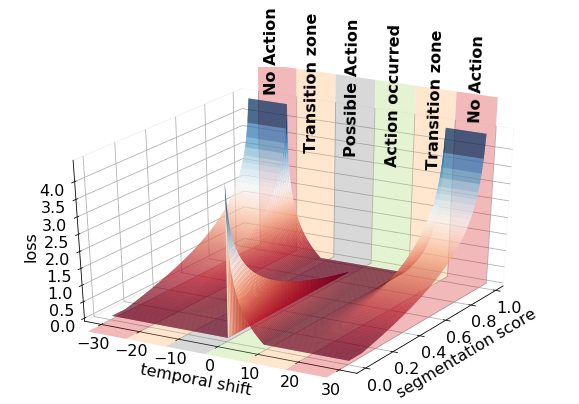}
    \caption{
    \textbf{Context-aware loss function.}
    %We design a novel loss that leverages the temporal context around an action spot. We heavily penalize the frames {\color{myred}\textbf{far-distant}} from the action and steadily decrease the penalty for {\color{myorange}\textbf{gradually closer}} frames. 
    %The loss does not penalize frames {\color{mygray}\textbf{just before}} the action to avoid providing misleading information as its occurrence remains uncertain, but heavily penalizes frames {\color{mygreen}\textbf{just after}} as the action occurred.
    We design a novel loss that leverages the temporal context around an action spot (at a temporal shift of $0$).
    We heavily penalize the frames {\color{myred}\textbf{far-distant}} from the action and decrease the penalty for those {\color{myorange}\textbf{gradually closer}}. 
    %First, we penalise the frames far distant from the action ({\color{myred}\textbf{red}}) and gradually decrease the penalty near the action ({\color{myorange}\textbf{orange}}).
    We do not penalize the frames {\color{mygray}\textbf{just before}} the action to avoid providing misleading information as its occurrence is uncertain, but we heavily penalize those {\color{mygreen}\textbf{just after}}, as the action has occurred.
    % we know for sure that the action has occurred.
    % The frames {\color{mygray}\textbf{just before}} the action are not penalized to avoid providing misleading information as its occurrence is uncertain. However, the loss heavaly penalize the frames {\color{mygreen}\textbf{just after}} the action occurred, as we know for sure that the action has occurred.
    % We design a novel loss that leverages the temporal context around an action spot. We heavily penalize the frames {\color{myred}\textbf{far-distant}} from the action and steadily decrease the penalty for the frames {\color{myorange}\textbf{gradually closer}} to the action. The frames {\color{mygray}\textbf{just before}} the action are not penalized to avoid providing misleading information as its occurrence is uncertain. However, those {\color{mygreen}\textbf{just after}} the action are heavily penalized as we know for sure that the action has occurred.
    }
    \label{fig:Loss}
\end{figure}

%% GENERAL INTRODUCTION ON SPORTS AND CV APPLICATION
Aside from automotive, consumer, and robotics applications, sports is considered one of the most valuable applications in computer vision~\cite{CVApplication}, capping \$91 billion of annual market revenue~\cite{GlobalSportsMarket}, with \$28.7 billion from the European Soccer market alone~\cite{EuropeanFootballMarket}. Recent advances helped provide automated tools to understand and analyze broadcast games. For instance, current computer vision methods can localize the field and its lines~\cite{farin2003robust,homayounfar2017sports}, detect players~\cite{Cioppa_2019_CVPR_Workshops,yang2017robust}, their motion~\cite{felsen2017will,manafifard2017survey}, their pose~\cite{Bridgeman_2019_CVPR_Workshops, Zecha_2019_CVPR_Workshops}, their team~\cite{Istasse_2019_CVPR_Workshops}, track the ball position~\cite{Sarkar_2019_CVPR_Workshops,Theagarajan_2018_CVPR_Workshops} and the camera motion~\cite{Lu2019Pan}. 
Understanding spatial frame-wise information is useful to enhance the visual experience of sports viewers~\cite{Rematas_2018_CVPR} and to gather players statistics~\cite{thomas2017computer}, but 
it misses higher-level game understanding.
% a higher-level game understanding is missing.
% Understanding such information is useful to enhance the visual experience of sports viewers~\cite{Rematas_2018_CVPR} and to gather players statistics~\cite{thomas2017computer}.
% However, they only focus on spatial frame-wise information, gatheriging per-player statistics~\cite{thomas2017computer} rather than higher-level game understanding.
%
%
%% CV FOR VIDEO AND BROADCASTER 
For broadcast producers, it is of paramount importance to have a deeper understanding of the game actions. For instance, live broadcast production follows specific patterns when particular actions occur; sports live reporters comment on the game actions; and highlights producers generate short summaries by ranking the most representative actions within the game.  In order to automate these production tasks, computer vision methods should understand the salient actions of a game and respond accordingly. While spatial information is widely studied and quite mature, 
%(as evidenced by current player and ball detectors)      
localizing actions in time remains a challenging task for current video understanding algorithms.

%% WHAT ARE WE PROPOSING IN THIS PAPER
In this paper, we target the action spotting challenge, with a primary application on soccer videos. The task of action spotting has been defined as the temporal localization of human-induced events annotated with a single timestamp~\cite{Giancola_2018_CVPR_Workshops}. Inherent difficulties arise from such annotations: their sparsity, the absence of start and end times of the actions, and their temporal discontinuities, \ie the unsettling fact that adjacent frames may be annotated differently albeit being possibly highly similar. To overcome these issues, we propose a novel loss that leverages the temporal context information naturally present around the actions, as depicted in Figure~\ref{fig:Loss}. To highlight its generality and versatility, we showcase how our loss can be used for the task of activity localization in ActivityNet~\cite{caba2015activitynet}, by spotting the beginning and end of each activity. Using the network BMN introduced in~\cite{Lin_2019_ICCV} and simply substituting their loss with our enhanced context-aware spotting loss function, we show an improvement of 0.15\% in activity proposal leading to a direct 0.38\% improvement in activity detection on ActivityNet~\cite{caba2015activitynet}. On the large-scale action spotting soccer-centric dataset, SoccerNet~\cite{Giancola_2018_CVPR_Workshops}, our network substantially increases the Average-mAP spotting metric from $49.7\%$ to $62.5\%$. %We will release the codes after the review process.

\mysection{Contributions.}
We summarize our contributions as follows. \textbf{(i)} We present a new loss function for temporal action segmentation further used for the task of action spotting, which is parameterized by the time-shifts of the frames from the ground-truth actions. \textbf{(ii)} We improve the performance of the state-of-the-art method on \ActivityNet by including our new contextual loss to detect activity boundaries, and improve the action spotting baseline of \SoccerNet by $12.8\%$. \textbf{(iii)} We provide detailed insights into our action spotting performance, as well as a qualitative application for automatic highlights generation.

\section{Related Work}
\label{sec:SOTA}

\mysection{Broadcast soccer video understanding.}
Computer vision tools are widely used in sports broadcast videos to provide soccer analytics~\cite{moeslund2014computer,thomas2017computer}. Current challenges lie in understanding high-level game information to identify salient game actions \cite{Cioppa_2018_CVPR_Workshops,Tsunoda_2017_CVPR_Workshops}, perform automatic game summarization \cite{sanabria2019deep,Shukla_2018_CVPR_Workshops,turchini2019flexible} and report commentaries of live actions~\cite{Yu_2018_CVPR}. Early work uses camera shots to segment broadcasts \cite{ekin2003automatic}, or analyze production patterns to identify salient moments of the game \cite{ren2005football}. Further developments have used low-level semantic information in Bayesian frameworks~\cite{huang2006semantic,tavassolipour2014event} to automatically detect salient game actions.

%ML with features + aggregator
Machine learning-based methods have been proposed to aggregate temporally hand-crafted features~\cite{baccouche2010action} or deep frame features \cite{jiang2016automatic} into recurrent networks~\cite{ramanathan2016detecting}. SoccerNet~\cite{Giancola_2018_CVPR_Workshops} provides an in-depth analysis of deep frame feature extraction and aggregation for action spotting in soccer game broadcasts. Multi-stream networks merge additional optical flow~\cite{Cai_2019_CVPR_Workshops,tsagkatakis2017goal} or excitement~\cite{bettadapura2016leveraging,Shukla_2018_CVPR_Workshops} information to improve game highlights identification. Furthermore, attention models are fed with per-frame semantic information such as pixel information~\cite{Cioppa_2018_CVPR_Workshops} or player localization~\cite{khan2018soccer} to extract targeted frame features. In our work, we leverage the temporal context information around actions to handle the intrinsic temporal patterns  representing these actions.

%% Extra on Dataset
Deep video understanding models are trained with large-scale datasets. While early works leveraged small custom video sets, a few large-scale datasets are available and worth mentioning, in particular Sports-1M~\cite{KarpathyCVPR14} for generic sports video classification, MLB-Youtube~\cite{Piergiovanni_2018_CVPR_Workshops} for baseball activity recognition, and GolfDB~\cite{McNally_2019_CVPR_Workshops} for golf swing sequencing. These datasets all tackle specific tasks in sports. In our work, we use SoccerNet~\cite{Giancola_2018_CVPR_Workshops} to assess the performance of our context-aware loss for action spotting in soccer videos.

\mysection{Video understanding.}
Recent video challenges~\cite{caba2015activitynet} include activity localization, that find temporal boundaries of activities.
Following object localization, two-stage approaches have been proposed including proposal generation~\cite{buch2017sst} and classification~\cite{sstad_buch_bmvc17}. SSN~\cite{zhao2017temporal} models each action instance with a structured temporal pyramid and TURN~TAP~\cite{gao2017turn} predicts action proposals and regresses the temporal boundaries, while GTAN~\cite{Long_2019_CVPR} dynamically optimizes the temporal scale of each action proposal with Gaussian kernels. BSN~\cite{Lin_2018_ECCV}, MGG~\cite{liu2019multi} and BMN~\cite{Lin_2019_ICCV} regress the time of activity boundaries, showing state-of-the-art performances on both ActivityNet 1.3~\cite{caba2015activitynet} and Thumos'14~\cite{THUMOS14}. Alternatively, ActionSearch~\cite{alwassel_2018_actionsearch} tackles the spotting task iteratively, learning to predict which frame to visit next in order to spot a given activity. However, this method requires sequences of temporal annotations by human annotators to train the models that are not readily available for datasets outside ActivityNet. Also, Alwassel~\etal~\cite{alwassel2018diagnosing} define an action spot as positive as soon as it lands within the boundary of an activity, which is less constraining than the action spotting defined in \SoccerNet.

Recently, Sigurdsson~\etal~\cite{sigurdsson2017actions} question boundaries sharpness and show that human agreement on temporal boundaries reach an average tIoU of $72.5\%$ for Charades~\cite{sigurdsson2016hollywood} and $58.7\%$ on MultiTHUMOS\cite{yeung2018every}. Alwassel~\etal~\cite{alwassel2018diagnosing} confirm such disparity on ActivityNet~\cite{caba2015activitynet}, but also show that it does not constitute a major roadblock to progress in the field. Different from activity localization, SoccerNet~\cite{Giancola_2018_CVPR_Workshops} proposes an alternative action spotting task for soccer action understanding, leveraging a well-defined set of soccer rules that define a single temporal anchor per action. In our work, we improve the \SoccerNet action spotting baseline by introducing a novel context-aware loss that temporally slices the vicinity of the action spots.
Also, we integrate our loss for generic activity localization and detection on a boundary-based method~\cite{Lin_2019_ICCV,Lin_2018_ECCV}.
\begin{figure*}[t]
    \centering
    \includegraphics[width=0.95\textwidth]{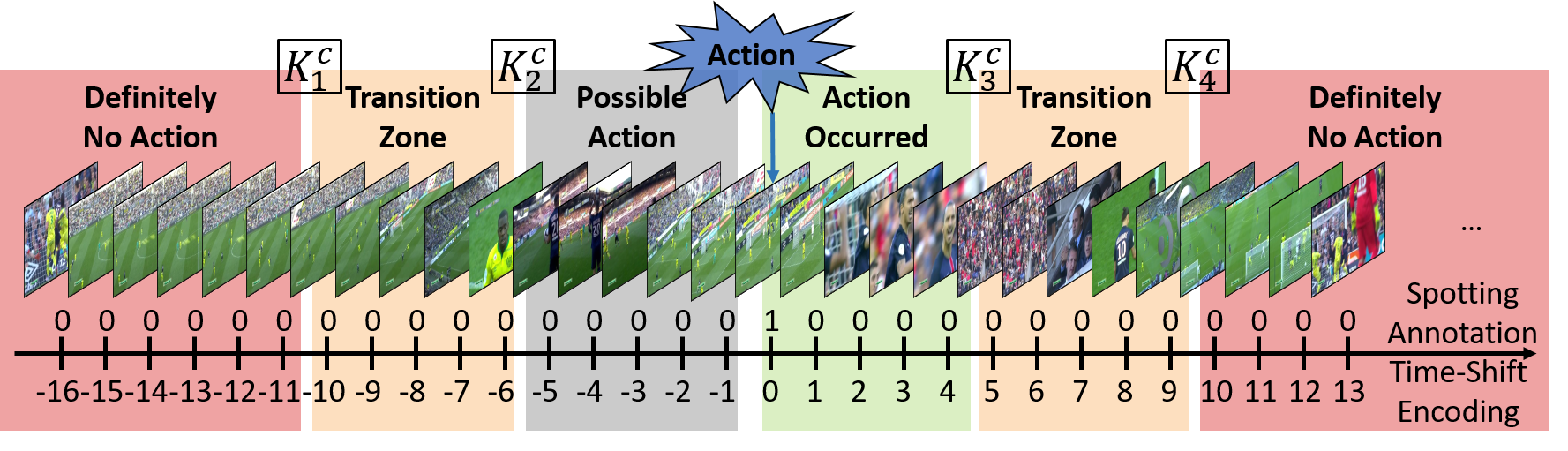}
    \caption{\textbf{Action context slicing.}
    We define six temporal segments around each ground-truth action spot, each of which induces a specific behavior in our context-aware loss function when training the network. {\color{myred}\textbf{Far before}} and {\color{myred}\textbf{far after}} the action, its influence is negligible, thus we train the network not to predict an action. {\color{mygray}\textbf{Just before}} the action, we do not influence the network since a particular context may or may not result in an action (\ie an attacking phase \textit{may} lead to a goal). {\color{mygreen}\textbf{Just after}} the action, its contextual information is rich and unambiguous as the action has just occurred (\ie a goal \textit{leads} to celebrating). Hence, we train the network to predict an action. Finally, we define {\color{myorange}\textbf{transition zones}} for our loss function to be smooth, in which we softly train the network not to predict an action. For each class $c$, the temporal segments are delimited by specific slicing parameters $K_i^c$ and are materialized through our time-shift encoding, which contains richer temporal context information about the action than the initial binary spotting annotation.
    }
    \label{fig:ActionSequencing}
\end{figure*}

\section{Methodology}
\label{sec:Method}
We address the action spotting task by developing a context-aware loss for a temporal segmentation module, and a YOLO-like loss for an action spotting module that outputs the \emph{spotting predictions} of the network. We first present the re-encoding of the annotations needed for the segmentation and spotting tasks,
then we explain how the losses of these modules are computed based on the re-encodings.

\mysection{Problem definition.}
We denote by $C$ the number of classes of the action spotting problem. Each action is identified by a single \emph{action frame} annotated as such. Each frame of a given video is annotated with either a one-hot encoded vector with $C$ components for the action frames or a vector of $C$ zeros for the background frames. We denote by $N_F$ the number of frames in a video.

\subsection{Encoding}
To train our network, the initial annotations are re-encoded in two different ways: with a \emph{time-shift encoding} used for the temporal segmentation loss, and with a \emph{YOLO-like encoding} used for the action spotting loss.

\mysection{Time-shift encoding (TSE) for temporal segmentation.}
We slice the temporal context around each action into segments related to their distance from the action, as shown in Figure~\ref{fig:ActionSequencing}. The segments regroup frames that are either \emph{far before}, \emph{just before}, \emph{just after}, \emph{far after} an action, or in \emph{transition zones} between these segments.

We use the segments in our temporal segmentation module so that its \emph{segmentation scores} reflect the following ideas. \textbf{(1)} \emph{Far before} an action spot of some class, we cannot foresee its occurrence. Hence, the score for that class should indicate that no action is occurring. \textbf{(2)} \emph{Just before} an action, its occurrence is uncertain. Therefore, we do not influence the score towards any particular direction. \textbf{(3)} \emph{Just after} an action has happened, plenty of visual cues allow for the detection of the occurrence of the action. The score for its class should reflect the presence of the action. \textbf{(4)} \emph{Far after} an action, the score for its class should indicate that it is not occurring anymore. The segments around the actions of class $c$ are delimited by four temporal \emph{context slicing parameters} $K^c_1 < K^c_2 < 0 < K^c_3 < K^c_4$ as shown in Figure~\ref{fig:ActionSequencing}.

The context slicing is used to perform a \emph{time-shift encoding} (TSE) of each frame $x$ of a video with a vector of length $C$, containing class-wise information on the relative location of $x$ with respect to its closest past or future actions. The TSE of $x$ for class $c$, noted $s^c(x)$, is the time-shift (\ie difference in frame indices) of $x$ from either its closest past or future ground-truth action of class $c$, depending on which has the dominant influence on $x$. We set $s^c(x)$ as the time-shift from the past action if either \textbf{(i)} $x$ is just after the past action; or \textbf{(ii)} $x$ is in the transition zone after the past action, but is far before the future action; or \textbf{(iii)} $x$ is in the transition zones after the past and before the future actions while being closer to the past action. In all other cases, $s^c(x)$ is the time-shift from the future action.

If $x$ is both located \emph{far after} the past action and \emph{far before} the future action, selecting either of the two time-shifts has the same effect in our loss. Furthermore, for the frames located either before the first or after the last annotated action of class $c$, only one time-shift can be computed and is thus set as $s^c(x)$. Finally, if no action of class $c$ is present in the video, then we set $s^c(x)=K^c_1$ for all the frames. This induces the same behavior in our loss as if they were all located far before their closest future action.

\mysection{YOLO-like encoding for action spotting.}
Inspired by YOLO~\cite{Redmon2016YOLO}, each ground-truth action of the video engenders an \emph{action vector} composed of $2+C$ values. The first value is a binary indicator of the presence ($=1$) of the action. The second value is the location of the frame annotated as the action, computed as the index of that frame divided by $N_F$. The remaining $C$ values represent the one-hot encoding of the action. We encode a whole video containing $N_\text{GT}$ actions in a matrix $\textbf{Y}$ of dimension $N_\text{GT}\times(2+C)$, with each line representing an action vector of the video.

\begin{figure}
    \centering
    \includegraphics[width=\linewidth]{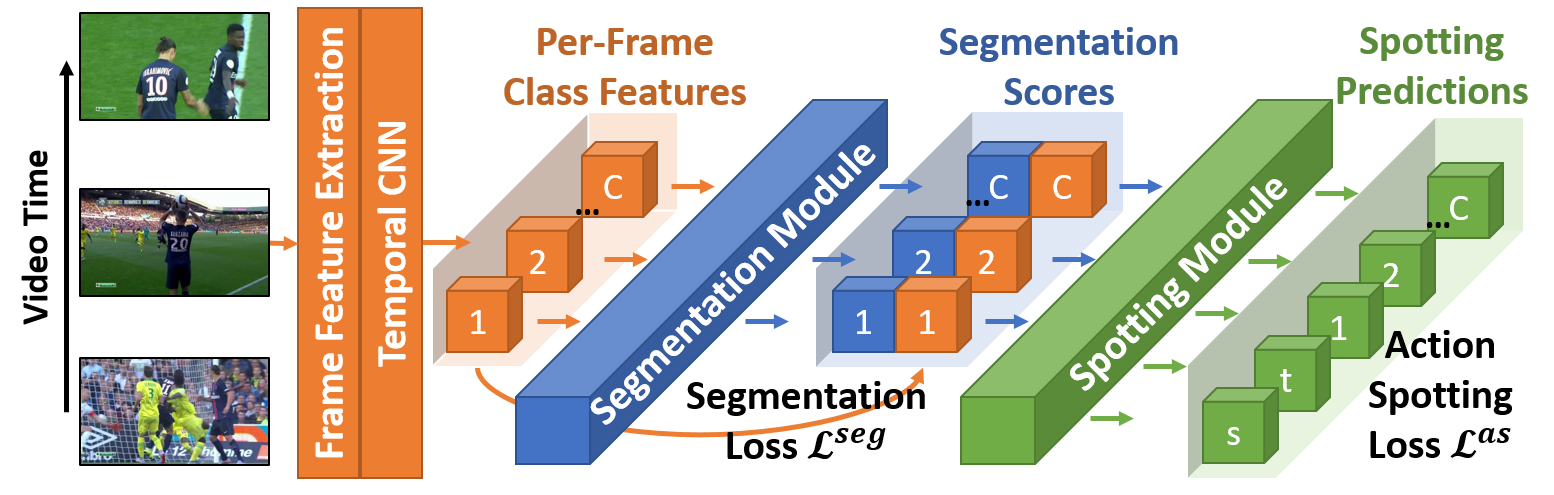}
    \caption{
    \textbf{Pipeline for action spotting.}
    We propose a network made of
    a \textbf{\color{Orange}frame feature extractor} and a \textbf{\color{Orange} temporal CNN} outputting $C$ class feature vectors per frame,
    a \textbf{\color{NavyBlue}segmentation module} outputting per-class segmentation scores, and 
    a \textbf{\color{Green}spotting module} extracting $2+C$ values per spotting prediction (\ie the confidence score $s$ for the spotting, its location $t$ and a per-class prediction).
    }
    \label{fig:Network}
\end{figure}

\subsection{Loss and Network Design}

\mysection{Temporal segmentation loss.} 
The TSE parameterizes the temporal segmentation loss described below. For clarity, we denote by $p$ the segmentation score for a frame $x$ to belong to class $c$ output by the segmentation module, and $s$ as the TSE of $x$ for class $c$. We detail the loss generated by $p$ in this setting, noted $L(p,s)$. First, in accordance with Figure~\ref{fig:ActionSequencing}, we compute $L(p,s)$ as follows:

\begin{numcases}{\hspace{-0.5cm}L(p,s)=}
   -\ln(1-p) & \hspace{-0.5cm}$s \leq K^c_1$ \label{eq:case1}
   \\
   -\ln\left(1-\frac{K^c_2-s}{K^c_2-K^c_1}\ p\right) & \hspace{-0.5cm}$K^c_1 < s \leq K^c_2$ \label{eq:case2}
   \\
   0 & \hspace{-0.5cm}$K^c_2 < s < 0$ \label{eq:case3}
   \\
   -\ln\left(\frac{s}{K^c_3} + \frac{K^c_3-s}{K^c_3}\ p\right) & \hspace{-0.5cm}$0 \leq s < K^c_3$ \label{eq:case4}
   \\
   -\ln\left(1-\frac{s-K^c_3}{K^c_4-K^c_3}\ p\right) & \hspace{-0.5cm}$K^c_3 \leq s < K^c_4$ \label{eq:case5}
   \\
   -\ln(1-p) & \hspace{-0.5cm}$s \geq K^c_4$. \label{eq:case6}
\end{numcases}

Then, following the practice in~\cite{Deliege2018HitNet,Sabour2017Dynamic} to help the network focus on improving its worst segmentation scores, we zero out the loss for scores that are satisfying enough. In the case of \Equation{case4} when $s=0$, we say that a score is satisfactory when it exceeds some \emph{maximum margin} $\tau_\text{max}$. In the cases of \Equations{case1}{case6}, we say that a score is satisfactory when it is lower than some \emph{minimum margin} $\tau_\text{min}$. The range of values for $p$ that leads to zeroing out the loss varies with $s$ and the slicing parameters in most cases. This is achieved by revising $L(p,s)$ as in \Equations{taumax}{taumin}. Figure~\ref{fig:Loss} shows a representation of $\tilde{L}(p,s)$.

\begin{numcases}{\hspace{-0.5cm}\tilde{L}(p,s)=}
    \hspace{-0.2cm}\max(0, L(p,s) + \ln(\tau_\text{max})) & \hspace{-0.9cm}$0 \leq s < K^c_3$ \label{eq:taumax}
   \\
   \hspace{-0.2cm}\max(0, L(p,s) + \ln(1-\tau_\text{min})) & \hspace{-0.6cm}otherwise. \label{eq:taumin}
\end{numcases}

Finally, the segmentation loss $\mathcal{L}^\text{seg}$ for a given video of frames $x_1,\ldots, x_{N_F}$ is given in \Equation{seg_loss}.

\begin{equation}\label{eq:seg_loss}
    \mathcal{L}^\text{seg} = \frac{1}{C \ N_F}\sum_{i = 1}^{N_F} \sum_{c=1}^C \tilde{L}(p^c(x_i),s^c(x_i))
\end{equation}

\mysection{Action spotting loss.} 
Let $N_\text{pred}$ be a fixed number of action spotting predictions generated by our network for each video. Those predictions are encoded in $\hat{\textbf{Y}}$ of dimension $N_\text{pred} \times (2+C)$, similarly to $\textbf{Y}$.

We leverage an iterative one-to-one matching algorithm to pair each of the $N_\text{GT}$ ground-truth actions with a prediction. First, we match each ground-truth location of $\textbf{Y}_{\cdot,2}$ with its closest predicted location in $\hat{\textbf{Y}}_{\cdot,2}$, and vice-versa (\ie we match the predicted locations with their closest ground-truth locations). Next, we form pairs of $(\text{ground-truth}, \text{predicted})$ locations that reciprocally match, we remove them from the process, and we iterate until all ground truths are coupled with a prediction. Consequently, we build $\hat{\textbf{Y}}^M$ as a reorganized version of the actions encoded in $\hat{\textbf{Y}}$, such that $\textbf{Y}_{i,2}$ and $\hat{\textbf{Y}}^M_{i,2}$ reciprocally match for all $i\leq N_\text{GT}$.

We define the action spotting loss $\mathcal{L}^\text{as}$ in \Equation{LossActionSpotting}. It corresponds to a weighted sum of the squared errors between the matched predictions and a regularization on the confidence score of the unmatched predictions.
\begin{equation}
    \hspace{-0.5mm}\mathcal{L}^\text{as} = \sum_{i=1}^{N_\text{GT}}\sum_{j=1}^{2+C}\alpha_j \left(\textbf{Y}_{i,j}-\hat{\textbf{Y}}^M_{i,j}\right)^2 \hspace{-1mm}+ \beta \hspace{-2mm}\sum_{i=N_\text{GT}+1}^{N_\text{pred}} \hspace{-1mm}\left(\hat{\textbf{Y}}^M_{i,1}\right)^2
    \label{eq:LossActionSpotting}
\end{equation}

\mysection{Complete loss.} 
The final loss $\mathcal{L}$ is presented in \Equation{LossFinal} as a weighted sum of $\mathcal{L}^\text{seg}$ and $\mathcal{L}^\text{as}$.

\begin{equation}
    \mathcal{L} = %\lambda^\text{as}\
    \mathcal{L}^\text{as} + \lambda^\text{seg}\ \mathcal{L}^\text{seg}
    \label{eq:LossFinal}
\end{equation}

\mysection{Network for action spotting.} 
The architecture of the network is illustrated in \Figure{Network} and further detailed in the \textbf{supplementary material}. We leverage frame feature representations for the videos (\eg ResNet) provided with the dataset, embodied as the output of the frame feature extractor of \Figure{Network}. The temporal CNN of \Figure{Network} is composed of a spatial two-layer MLP, followed by four multi-scale 3D convolutions (\ie across time, features and classes). The temporal CNN outputs a set of $C \times f$ features for each frame organized in $C$ feature vectors (one per class) of size $f$, as in~\cite{Sabour2017Dynamic}. These features are input into a segmentation module, in which we use Batch Normalization~\cite{Ioffe2015BatchNorm} and sigmoid activations. The closeness of the $C$ vectors obtained in this way to a pre-defined vector gives the $C$ segmentation scores output by the segmentation module, as~\cite{Deliege2018HitNet}. The $C \times f$ features obtained previously are concatenated with the $C$ scores and fed to the action spotting module, as shown in Figure~\ref{fig:Network}. It is composed of three successive temporal max-pooling and 3D convolutions, and outputs $N_\text{pred}$ vectors of dimension $(2+C)$. The first two elements of these vectors are sigmoid-activated, the $C$ last are softmax-activated. The activated vectors are stacked to produce the prediction $\hat{\textbf{Y}}$ of dimension $N_\text{pred} \times (2+C)$ for the action spotting task.

\section{Experiments}
\label{sec:Exp}

We evaluate our new context-aware loss function in two scenarios: the action spotting task of \SoccerNet, and activity localization and detection tasks on ActivityNet~\cite{caba2015activitynet}.

\subsection{Experiments on SoccerNet}

\mysection{Data.} Three classes of action are annotated in SoccerNet by Giancola~\etal~\cite{Giancola_2018_CVPR_Workshops}: goals, cards, and substitutions, so $C=3$ in this case. They identify each action by one annotated frame: the moment the ball crosses the line for \emph{goal}, the moment the referee shows a player a card for \emph{card}, and the moment a new player enters the field for \emph{substitution}. We train our network on the frame features already provided with the dataset. Giancola~\etal first subsampled the raw videos at $2$ fps, then they extracted the features with a backbone network and reduced them by PCA to $512$ features for each frame of the subsampled videos. Three sets of features are provided, each extracted with a particular backbone network: I3D~\cite{Carreira_2017_CVPR}, C3D~\cite{Tran2015ICCV}, and ResNet~\cite{He_2016_CVPR}.

\mysection{Action spotting metric.} 
We measure performances with the action spotting metric introduced in SoccerNet~\cite{Giancola_2018_CVPR_Workshops}. An action spot is defined as positive if its temporal offset from its closest ground truth is less than a given tolerance $\delta$. The average precision (AP) is estimated based on Precision-Recall curves, then averaged between classes (mAP). An Average-mAP is proposed as the AUC of the mAP over different tolerances $\delta$ ranging from 5 to 60 seconds.

\mysection{Experimental setup.} We train our network on batches of \emph{chunks}. We define a chunk as a set of $N_F$ contiguous frame feature vectors. We set $N_F=240$ to maintain a high training speed while retaining sufficient contextual information. This size corresponds to a clip of $2$ minutes of raw video. A batch contains chunks extracted from a single raw video. We extract a chunk around each ground-truth action, such that the action is randomly located within the chunk. Then, to balance the batch, we randomly extract $N_\text{GT}/C$ chunks composed of background frames only. An epoch ends when the network has been trained on one batch per training video. At each epoch, new batches are re-computed for each video for data augmentation purposes. Each raw video is time-shift encoded before training. Each new training chunk is encoded with the YOLO-like encoding.

The number of action spotting predictions generated by the network is set to $N_\text{pred}=5$, as we observed that no chunks of $2$ minutes of raw video contain more than $5$ actions. We train the network during $1000$ epochs, with an initial learning rate $lr=10^{-3}$ linearly decreasing to $10^{-6}$. We use Adam as the optimizer with default parameters~\cite{Diederick2015Adam}.

%Losses
For the segmentation loss, we set the margins $\tau_\text{max}=0.9$ and $\tau_\text{min}=0.1$ in \Equations{taumax}{taumin}, following the practice in~\cite{Sabour2017Dynamic}. For the action spotting loss in \Equation{LossActionSpotting}, we set $\alpha_j=1$ for $j\neq 2$, while $\alpha_2$ is optimized (see below) to find an appropriate weighting for the location components of the predictions. Similarly, $\beta$ is optimized to find the balance between the loss of the action vectors and the regularization of the remaining predictions. For the final loss in \Equation{LossFinal}, we optimize $\lambda^\text{seg}$ to find the balance between the two losses.

\mysection{Hyperparameter optimization.}
For each set of features (I3D, C3D, ResNet), we perform a joint Bayesian optimization~\cite{BayesianOpt} on the number of frame features $f$ extracted per class, on the temporal receptive field $r$ of the network (\ie temporal kernel dimension of the 3D convolutions), and on the parameters $\alpha_2, \beta, \lambda^\text{seg}$. Next, we perform a grid search optimization on the slicing parameters $K_i^c$. 

For ResNet, we obtain $f=16,\ r=80,\ \alpha_2=5,\ \beta=0.5,\ \lambda^\text{seg}=1.5$. For goals (resp. cards, substitutions) we have $K_1=-40$ (resp. $-40$, $-80$), $K_2=-20$ (resp. $-20$, $-40$), $K_3=120$ (resp. $20$, $20$), and $K_4=180$ (resp. $40$, $40$). Given the framerate of 2 fps, those values can be translated to seconds by scaling them down by a factor of 2. The value $r=80$ corresponds to a temporal receptive field of $20$ seconds on both sides of the central frame in the temporal dimension of the 3D convolutions.

\begin{table}[t]
\begin{centering}
\begin{tabular}{c||c|c|c}
\multirow{2}{*}{Method} & \multicolumn{3}{c}{Frame features}\tabularnewline
 & I3D & C3D & ResNet\tabularnewline\hline\hline
SoccerNet baseline 5s~\cite{Giancola_2018_CVPR_Workshops} & - & - & 34.5\tabularnewline\hline 
SoccerNet baseline 60s~\cite{Giancola_2018_CVPR_Workshops} & - & - & 40.6\tabularnewline\hline 
SoccerNet baseline 20s~\cite{Giancola_2018_CVPR_Workshops} & - & - & 49.7\tabularnewline\hline\hline
Vats \etal~\cite{Vats2019Event_full} & - & - & 57.5\tabularnewline\hline\hline
Ours & 53.6 & 57.7 & \textbf{62.5}\tabularnewline
\end{tabular}
\caption{\textbf{Results on SoccerNet.} Average-mAP (in \%) on the test set of SoccerNet for the action spotting task. We establish a new state-of-the-art performance.}
\label{tab:results}
\par\end{centering}
\end{table}

\mysection{Main results.} 
The performances obtained with the optimized parameters are reported in Table~\ref{tab:results}. As shown, we establish a new state-of-the-art performance on the action spotting task of SoccerNet, outperforming the previous benchmark by a comfortable margin, for all the frame features. ResNet gives the best performance, as also observed in~\cite{Giancola_2018_CVPR_Workshops}. A sensitivity analysis of the parameters $K_i^c$ reveals robust performances around the optimal values, indicating that no heavy fine-tuning is required for the context slicing. Also, performances largely decrease as the slicing is strongly reduced, which emphasizes its usefulness.

\mysection{Ablation study.} 
Since the ResNet features provide the best performance, we use them with their optimized parameters for the following ablation studies. \textbf{(i)} We remove the segmentation module, which is equivalent to setting $\lambda^\text{seg}=0$ in \Equation{LossFinal}. This also removes the context slicing and the margins $\tau_\text{max}$ and $\tau_\text{min}$. \textbf{(ii)} We remove the action context slicing such that the ground truth for the segmentation module is the raw binary annotations, \ie all the frames must be classified as background except the action frames. This is equivalent to setting $K_1=-1=K_2=-K_3=-K_4$. \textbf{(iii)} We remove the margins that help the network focus on improving its worst segmentation scores, by setting $\tau_\text{max}=1,\ \tau_\text{min}=0$ in \Equations{taumax}{taumin}. \textbf{(iv)} We remove the iterative one-to-one matching between the ground truth $\textbf{Y}$ and the predictions $\hat{\textbf{Y}}$ before the action spotting loss, which is equivalent to using $\hat{\textbf{Y}}$ instead of $\hat{\textbf{Y}}^M$ in \Equation{LossActionSpotting}. The results of the ablation studies are shown in \Table{ablation}.

From an Average-mAP perspective, the auxiliary task of temporal segmentation improves the performance on the action spotting task (from $58.9\%$ to $62.5\%$), which is a common observation in multi-task learning~\cite{Zamir_2018_CVPR}. When the segmentation is performed, our temporal context slicing gives a significant boost compared to using the raw binary annotations (from $57.8\%$ to $62.5\%$). This observation is in accordance with the sensitivity analysis. It also appears that it is preferable to not use the segmentation at all rather than using the segmentation with the raw binary annotations ($58.9\%$ vs $57.8\%$), which further underlines the usefulness of the context slicing. A boost in performance is also observed when we use the margins to help the network focus on improving its worst segmentation scores (from $59.0\%$ to $62.5\%$). Eventually, Table~\ref{tab:ablation} shows that it is extremely beneficial to match the predictions of the network with the ground truth before the action spotting loss (from $46.8\%$ to $62.5\%$). This makes sense since there is no point in evaluating the network on its ability to order its predictions, which is a hard and unnecessary constraint. The large impact of the matching is also justified by its direct implication in the action spotting task assessed through the Average-mAP.

\begin{table}
    \centering
    \begin{tabular}{c||c|c|c|c||c}
    & Segm. & Slic. & Marg. & Match. & Result\\ \hline \hline 
    (i)   &  &  &  & \checkmark & 58.9\\ \hline 
    (ii)  & \checkmark &  & \checkmark & \checkmark & 57.8\\ \hline 
    (iii) & \checkmark & \checkmark &  & \checkmark & 59.0\\ \hline 
    (iv)  & \checkmark & \checkmark & \checkmark &  & 46.8\\ \hline\hline
    Ours  & \checkmark & \checkmark & \checkmark & \checkmark & \textbf{62.5}\\ 
    \end{tabular}
    \caption{
    \textbf{Ablation study.} 
    We perform ablations by 
    \textbf{(i)} removing the segmentation ($\lambda^\text{seg}=0$), hence the slicing and the margins; 
    \textbf{(ii)} removing the context slicing ($K_1=-1=K_2=-K_3=-K_4$); 
    \textbf{(iii)} removing the margins that help the network focus on improving its worst segmentation scores ($\tau_\text{min}=0$, $\tau_\text{max}=1$); 
    \textbf{(iv)} removing the matching (using $\hat{\textbf{Y}}$ instead of $\hat{\textbf{Y}}^M$ in $\mathcal{L}^\text{as}$). Each part evidently contributes to the overall performance.
    }
    \label{tab:ablation}
\end{table}

\begin{figure}
    \centering
    \includegraphics[width=\linewidth]{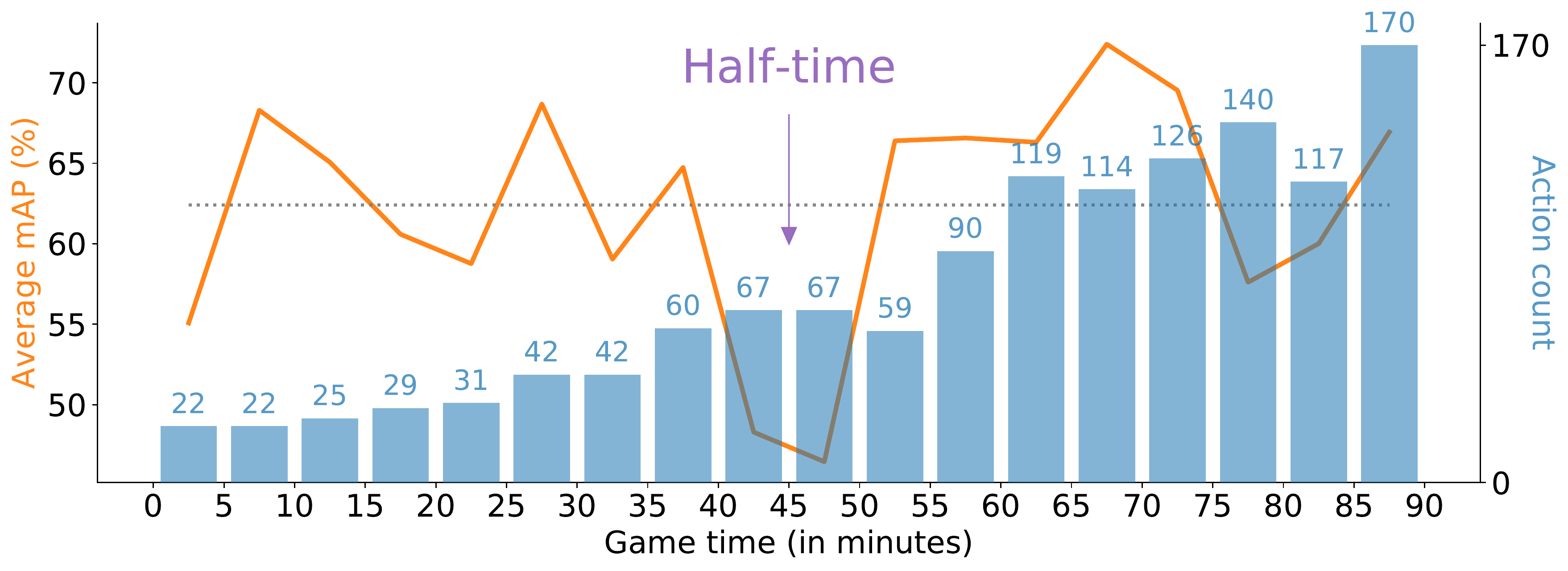}
    \caption{\textbf{Performance as function of game time.} 
    {\color{anthoorange}\textbf{Average-mAP}} spotting performance over the game time with all ground-truth actions of the dataset binned in $5$ minute intervals. 
    It appears that actions around the half-time break are more challenging to spot.
    {\color{anthoblue}\textbf{Number of actions}} for each bin.
    {\color{mygray}\textbf{Our performance ($62.5\%$)}}.}
    \label{fig:perfs-through-time}
\end{figure}

\mysection{Results through game time.}
In soccer, it makes sense to analyze the performance of our model through game time, since the actions are not uniformly distributed throughout the game. For example, a substitution is more likely to occur during the second half of a game. We consider non-overlapping bins corresponding to $5$ minutes of game time and compute the Average-mAP for each bin. \Figure{perfs-through-time} shows the evolution of this metric through game time.

It appears that actions occurring during the first five minutes of a half-time are substantially more difficult to spot than the others. This may be partially explained by the occurrence of some of these actions at the very beginning of a half-time, for which the temporal receptive field of the network requires the chunk to be temporally padded. Hence, some information may be missing to allow the network to spot those actions. Besides, when substitutions occur during the break, they are annotated as such on the first frame of the second halves of the matches, which makes them practically impossible to spot. In the test set, this happens for $28\%$ of the matches. None of these substitutions are spotted by our model, which thus degrades the performances during the first minutes of play in the second halves of the matches. However, they merely represent $5\%$ of all the substitutions, and removing them from the evaluation only boosts our Average-mAP by $0.7\%$ (from $62.5\%$ to $63.2\%$).

\begin{figure}
    \centering
    \includegraphics[width=\linewidth]{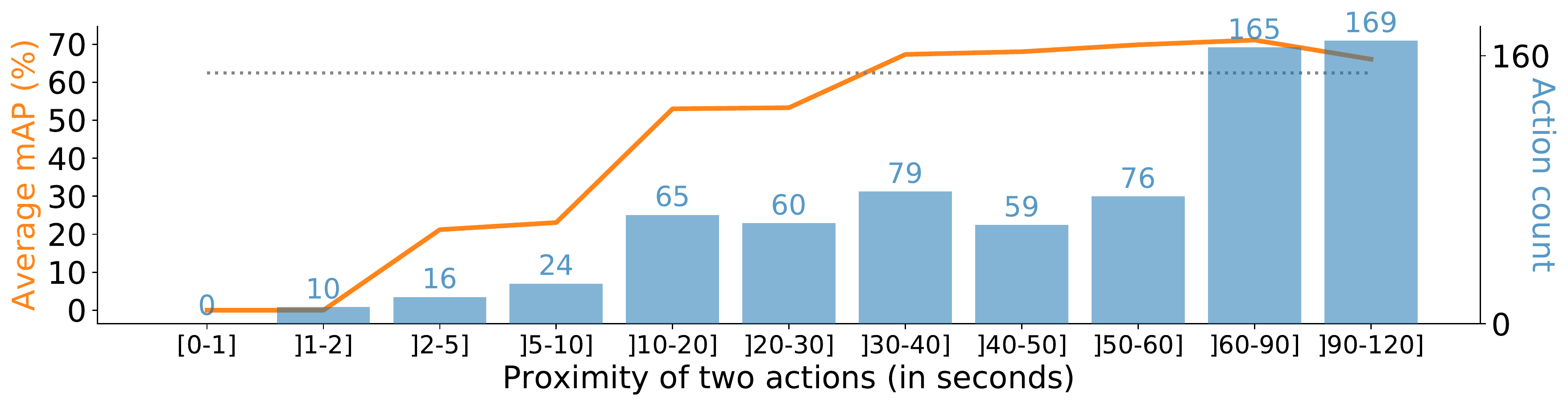}
    \caption{\textbf{Performance as function of action vicinity.} 
    {\color{anthoorange}\textbf{Average-mAP}} spotting performance per bin of ground-truth actions grouped by distance (in seconds) from their closest  ground-truth action.
    It appears that nearby actions are more challenging to spot.
    {\color{anthoblue}\textbf{Number of actions}} for each bin.
    {\color{mygray}\textbf{Our performance ($62.5\%$)}}.}
    \label{fig:perfs-closeness}
\end{figure}

\mysection{Results as function of action vicinity.}
We investigate whether actions are harder to spot when they are close to each other. We bin the ground-truth actions based on the distance that separates them from the previous (or next, depending on which is the closest) ground-truth action, regardless of their classes. Then, we compute the Average-mAP for each bin. The results are represented in Figure~\ref{fig:perfs-closeness}. 

We observe that the actions are more difficult to spot when they are close to each other. This could be due to the reduced number of visual cues, such as replays, when an action occurs rapidly after another and thus must be broadcast. Some confusion may also arise because the replays of the first action can still be shown after the second action, \eg a sanctioned foul followed by a converted penalty.
This analysis also shows that the action spotting problem is challenging even when the actions are further apart, as the performances in Figure~\ref{fig:perfs-closeness} eventually plateau.

\begin{figure}
    \centering
    \includegraphics[width=0.95\columnwidth]{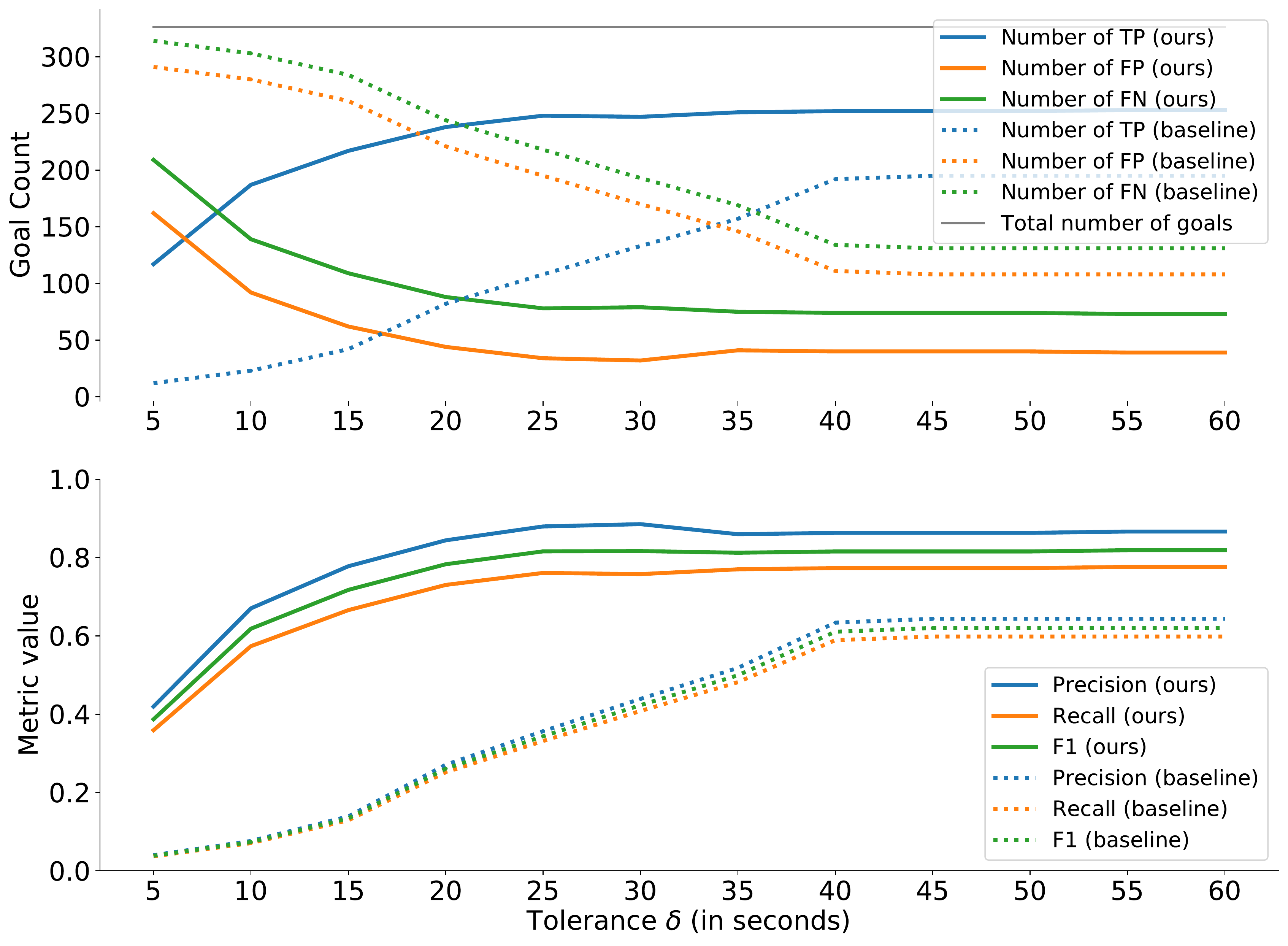}\\
    \caption{\textbf{Per-class results (goals).} A prediction of class \emph{goal} is a {\color{anthoblue}\textbf{true positive (TP)}} with tolerance $\delta$ when it is located at most $\delta/2$ seconds from a ground-truth goal. The baseline results are obtained from the best model of~\cite{Giancola_2018_CVPR_Workshops}. Our model spots most goals within $10$ seconds around the ground truth ($\delta=20$ seconds).
    }
    \label{fig:goalmetrics}
\end{figure}

\mysection{Per-class results.}
We perform a per-class analysis in a similar spirit as the Average-mAP metric. For a given class, we fix a tolerance $\delta$ around each annotated action to determine positive predictions and we aggregate these results in a confusion matrix. An action is considered spotted when its confidence score exceeds some threshold optimized for the $F_1$ score on the validation set. From the confusion matrix, we compute the precision, recall and $F_1$ score for that class and for that tolerance $\delta$. Varying $\delta$ from $5$ to $60$ seconds provides the evolution of the three metrics as a function of the tolerance. Figure~\ref{fig:goalmetrics} shows these curves for \emph{goals} for our model and for the baseline \cite{Giancola_2018_CVPR_Workshops}. The results for cards and substitutions are provided in \textbf{supplementary material}.

Figure~\ref{fig:goalmetrics} shows that most goals can be efficiently spotted by our model within $10$ seconds around the ground truth ($\delta=20$ seconds). We achieve a precision of $80\%$ for that tolerance. The previous baseline plateaus within $20$ seconds ($\delta=40$ seconds) and still has a lower performance. In particular for goals, many visual cues facilitate their spotting, \eg multiple replays, particular camera views, or celebrations from the players and from the public.

\subsection{Experiments on ActivityNet}

In this section, we evaluate our context-aware loss in a more generic task than action spotting in soccer videos. We tackle the \emph{Activity Proposal} and \emph{Activity Detection} tasks of the challenging ActivityNet dataset, for which we use the ResNet features provided with the dataset at $5$ fps.

\mysection{Setup.} 
We use the current state-of-the-art network, namely BMN \cite{Lin_2019_ICCV}, with the code provided in~\cite{BMNCode}. BMN is equipped with a temporal evaluation module (TEM), which plays a similar role as our temporal segmentation module. We replace the loss associated with the TEM by our novel temporal segmentation loss $\mathcal{L}^\text{seg}$. The slicing parameters are set identically for all the classes and are optimized with respect to the AUC performance on the validation set by grid search with the constraint $K_1 = 2 K_2 = -2 K_3 = -K_4$. The optimization yields the best results where $K_1 = -14$.

\mysection{Results.} 
The average performances on $20$ runs of our experiment and of the BMN base code~\cite{BMNCode} are reported in Table~\ref{tab:results-ActivityNet}. Our novel temporal segmentation loss improves the performance obtained with BMN~\cite{BMNCode} by $0.15\%$ and $0.12\%$ for the activity proposal task (AR@100 and AUC) and by $0.38\%$ for the activity detection task (Average-mAP). These increases compare with some recent increments, while being obtained just by replacing their TEM loss by our context-aware segmentation loss. The network thus has the same architecture and number of parameters. We conjecture that our loss $\mathcal{L}^\text{seg}$, through its particular context slicing, helps train the network by modelling the uncertainty surrounding the annotations. Indeed, it has been shown in \cite{alwassel2018diagnosing,sigurdsson2017actions} that a large variability exists among human annotators on which frames to annotate as the beginning and the end of the activities of the dataset. Let us note that in BMN, the TEM loss is somehow adapted around the action frames in order to mitigate the penalization attributed to their neighboring frames. Our work goes one step further, by directly designing a temporal context-aware segmentation loss.

\begin{table}
\begin{centering}
\begin{tabular}{l||c|c|c}
Method & AR@$100$ & AUC & Average-mAP \\\hline 
Lin~\etal~\cite{Lin2017Temporal}      & $73.01$ & $64.40$ & $29.17$\\\hline 
Gao~\etal~\cite{Gao_2018_ECCV}        & $73.17$ & $65.72$ & - \\\hline 
BSN~\cite{Lin_2018_ECCV}              & $74.16$ & $66.17$ & $30.03$\\\hline 
P-GCN~\cite{zeng2019graph_full}       & -       & -       & $31.11$\\\hline 
BMN~\cite{Lin_2019_ICCV}              & $75.01$ & $67.10$ & $\mathbf{33.85}$\\\hline\hline
BMN code~\cite{BMNCode}               & $75.11$ & $67.16$ & $30.67\pm0.08$\\\hline
Ours: \cite{BMNCode} + $\mathcal{L}^\text{seg}$ & $\mathbf{75.26}$ & $\mathbf{67.28}$ & $31.05\pm0.07$\\
\end{tabular}
\caption{
\textbf{Results on ActivityNet} validation set for the proposal task (AR@100, AUC) and for the detection task (Average-mAP). For our experiments, we report the average values on $20$ runs.}
%\textbf{Results on ActivityNet.} 
%The metrics are given for the validation set of ActivityNet for the proposal task (AR@100 and AUC) and for the detection task (Average-mAP). 
%For our experiments, we report the average values on $20$ runs.}
\label{tab:results-ActivityNet}
\par\end{centering}
\end{table}

\section{Automatic Highlights Generation for Soccer}
\label{sec:Discussion}

\begin{figure}
    \centering
    \includegraphics[width=0.45\textwidth]{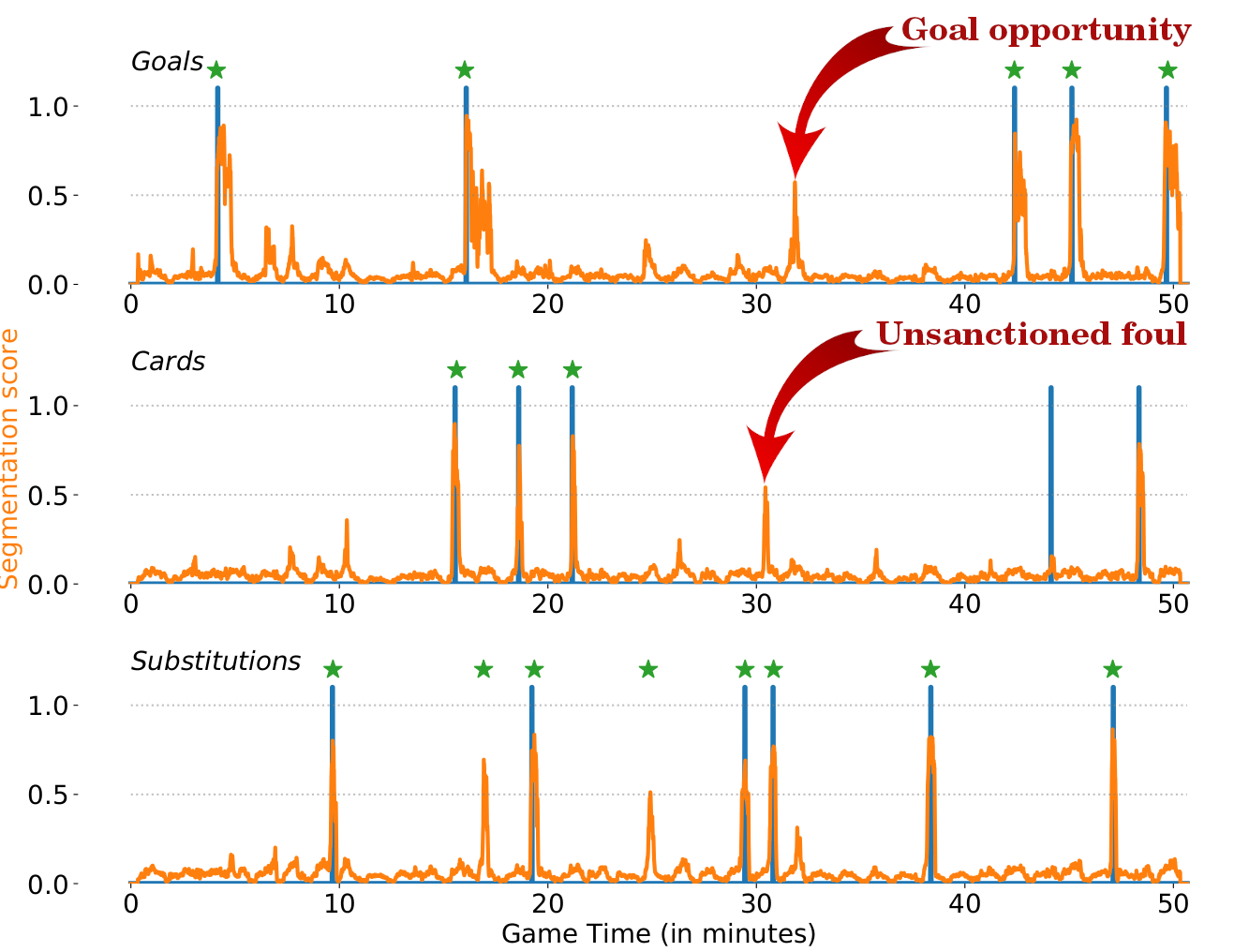}
%    \caption{\textbf{Action spotting and segmentation} for the second half of the famous ``Remuntada" match, Barcelona - PSG, in March 2017. {\color{anthoblue} \textbf{Ground truth actions}}, {\color{anthoorange}\textbf{temporal segmentation curves}}, and {\color{anthogreen}\textbf{spotting results (green stars)}} are illustrated. {\color{anthored}\textbf{Unannotated interesting actions}} can be identified using our segmentation. %\BG{make the font of the y axis label bigger}
    \caption{\textbf{Action spotting and segmentation} for the $2^\text{nd}$ half of the ``Remuntada" FCB - PSG. {\color{anthoblue} \textbf{Ground truth actions}}, {\color{anthoorange}\textbf{temporal segmentation curves}}, and {\color{anthogreen}\textbf{spotting results}} are illustrated.  We can identify {\color{anthored}\textbf{unannotated interesting actions}} using our segmentation. %\BG{make the font of the y axis label bigger}
    }
    \label{fig:predsandsegs}
\end{figure}

Some action spotting and temporal segmentation results are shown in Figure~\ref{fig:predsandsegs}. It appears that some sequences of play have a high segmentation score for some classes but do not lead, quite rightly, to an action spotting. It turns out that these sequences are often related to unannotated actions of supplementary classes that resemble those considered so far, such as \emph{unconverted goal opportunities} and \emph{unsanctioned fouls}. Video clips of the two actions identified in Figure~\ref{fig:predsandsegs} are provided in the \textbf{supplementary material}.

\begin{figure}
    \centering
    \includegraphics[width=\linewidth]{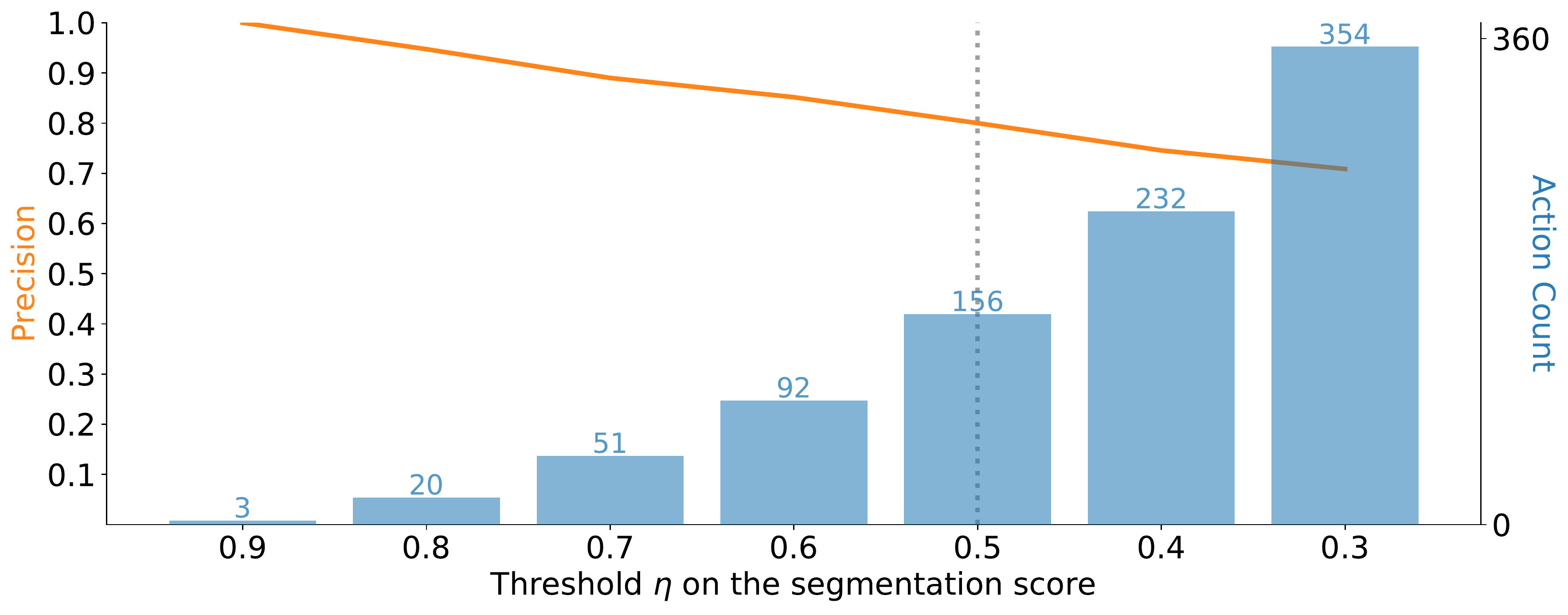}
    \caption{\textbf{Precision for goal opportunities}, as a function of the threshold on the segmentation score to exceed for manually inspecting a sequence. For scores larger than $\eta=0.5$, a {\color{anthoorange}\textbf{precision}} of $0.8$ is achieved, \ie $80\%$ of the sequences inspected were goal opportunities. {\color{anthoblue}\textbf{Number of sequences}} inspected per threshold.}
    \label{fig:PerformanceHighlight}
\end{figure}

To quantify the spotting results of goal opportunities, we can only compute the precision metric since these actions are not annotated. We manually inspect each video sequence of the test set where the segmentation score for goals exceeds some threshold $\eta$ but where no ground-truth goal is present. We decide whether the sequence is a goal opportunity or not by asking two frequent observers of soccer games if they would include it in the highlights of the match. The sequence is a true positive when they both agree to include it and a false positive, otherwise. The precision is then computed for that $\eta$. By gradually decreasing $\eta$ from $0.9$ to $0.3$, we obtain the precision curve shown in Figure~\ref{fig:PerformanceHighlight}. It appears that $80\%$ of the sequences with a segmentation score larger than $\eta=0.5$ are considered goal opportunities. 
%Also, the two observers disagreed with respect to what they considered to be an interesting sequence for only $4\%$ of the sequences, all of which having a low segmentation score. 

As a direct by-product, we %can 
derive an %a simple 
automatic highlights generator without explicit supervision. We extract a video clip starting $15$ seconds before each spotting of a \emph{goal} or a \emph{card} and ending $20$ seconds after. We proceed likewise for the sequences with a segmentation score $\geq 0.5$ for \emph{goals}. We dismiss substitutions as they rarely appear in highlights. %Substitutions are not considered here, since they almost never appear in highlights. 
We assemble the clips chronologically to produce the highlights video, provided in \textbf{supplementary material}. %as provided in the \textbf{supplementary material}. 
%The evaluation of 
Evaluating its %the %overall 
quality %of this video 
is subjective, but we found its content to be adequate, even if the montage could be improved. Indeed, only sequences where a goal, a goal opportunity, or a foul occurs are selected. This reinforces the usefulness of the segmentation,  %task, 
as it provides a direct overview of the proceedings of the match, including proposals for unannotated actions that are %usually 
interesting for highlights.

\section{Conclusion}
\label{sec:Conclusion}

We tackle the challenging action spotting task of SoccerNet with a novel context-aware loss for segmentation and a YOLO-like loss for the spotting. The former treats the frames according to their time-shift from their closest ground-truth actions. The latter leverages an iterative matching algorithm that alleviates the need for the network to order its predictions. To show generalization capabilities, we also test our context-aware loss on ActivityNet. We improve the state-of-the-art on ActivityNet by $0.15\%$ in AR@100, $0.12\%$ in AUC, and $0.38\%$ in Average-mAP, %by only  
by only including our context-aware loss without changing the network architecture. We achieve a new state-of-the art on SoccerNet, surpassing by far the previous baseline (from $49.7\%$ to $62.5\%$ in Average-mAP) and spotting most actions within $10$ seconds around their ground truth. %Both our context-aware loss and matching algorithm are shown to be key components in this performance. 
Finally, we leverage the resulting segmentation results to identify unannotated actions such as goal opportunities and derive a highlights generator without specific supervision.

\mysection{Acknowledgments.} This work is supported by the DeepSport project of the Walloon region and the FRIA (Belgium), as well as the King Abdullah University of Science and Technology (KAUST) Office of Sponsored Research (OSR) under Award No. OSR-CRG2017-3405.
\clearpage

{\small
\bibliographystyle{ieee_fullname}
\bibliography{egbib}
}

% Not for the main submission
% \newpage
\clearpage

\begin{figure*}
    \centering
    \includegraphics[width=\linewidth]{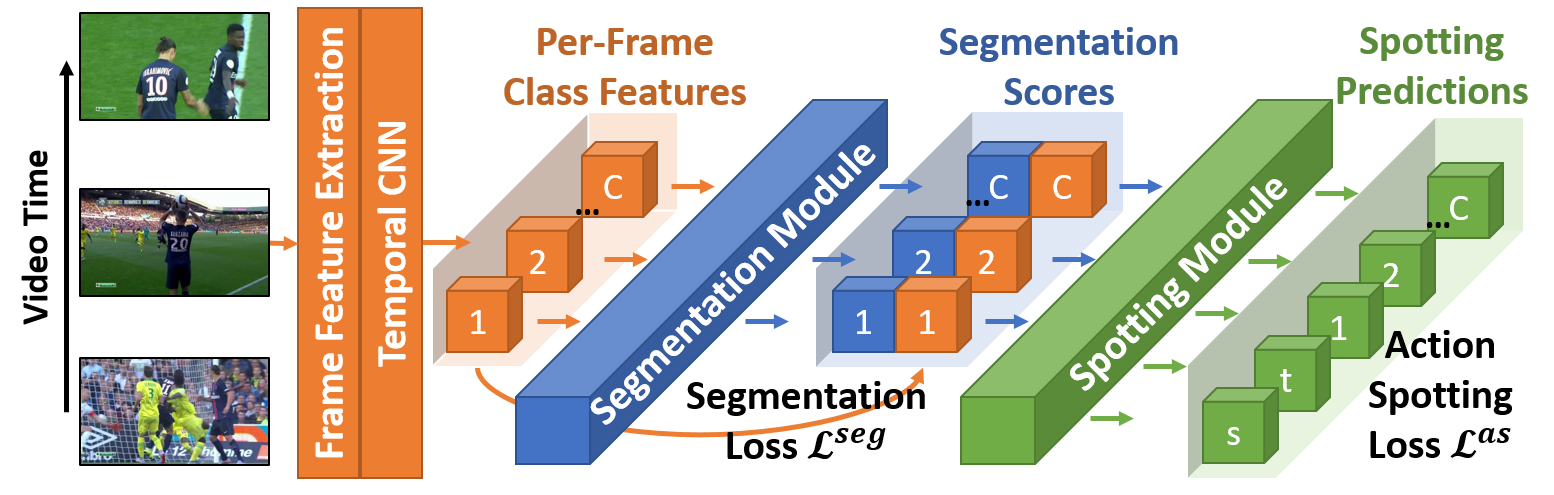}
    \caption{
    \textbf{Pipeline for action spotting.}
    We propose a network made of
    a \textbf{\color{Orange}frame feature extractor} and a \textbf{\color{Orange} temporal CNN} outputting $C$ class feature vectors per frame,
    a \textbf{\color{NavyBlue}segmentation module} outputting per-class segmentation scores, and 
    a \textbf{\color{Green}spotting module} extracting $2+C$ values per spotting prediction (\ie the confidence score $s$ for the spotting, its location $t$ and a per-class prediction).
    }
    \label{fig:Network_Supp}
\end{figure*}

% \newpage
% ~
% \newpage

\section{Supplementary Material}

A supplementary video summarizing our work is available on 
\url{https://youtu.be/FAeWxs0d4_o}.

\subsection{Notations}
Let us recall the following notations from the paper:
\begin{itemize}
    \item $C$ is the number of classes in the spotting task.
    \item $N_F$ is the number of frames in the chunk considered.
    \item $N_\text{GT}$ is the number of ground-truth actions in the chunk considered.
    \item $N_\text{pred}$ is the number of predictions output by the network for the spotting task.
    \item $f$ is the number of features computed for each class, for each frame, before the segmentation module (see Figure~\ref{fig:Network_Supp}).
    \item $r$ is the temporal receptive field of the network (used in the temporal convolutions).
    \item $\hat{\textbf{Y}}$ regroups the spotting predictions of the network, and has  dimension $N_\text{pred}\times (2+C)$. The first column represents the confidence scores for the spots, the second contains the predicted locations, and the other are per-class classification scores.
    \item $\textbf{Y}$ encodes the ground-truth action vectors of the chunk considered, and has dimension $N_\text{GT}\times (2+C)$. 
    \item $K^c_i$ ($i=1,2,3,4$) denotes the context slicing parameters of class $c$.
\end{itemize}

We also use the following notations for the layers of a convolutional neural network:
\begin{itemize}
    \item FC($n$) is a fully connected layer (\eg in a multi-layer perceptron) between any vector to a vector of size $n$. 
    \item ReLU is the rectified linear unit.
    \item Conv($n, p\times q$) is a convolutional layer with $n$ kernels of dimensions $p\times q$.
    %\item MaxPool($p\times q$) is a max-pooling layer that pools values in $p\times q$ pooling regions 
\end{itemize}

\subsection{Detailed Network Architecture for SoccerNet}

The architecture of the network used in the paper for the action spotting task of \SoccerNet, as depicted in Figure~\ref{fig:Network_Supp}, is detailed hereafter.

\textbf{\color{Orange}1. 
Frame feature extractor and temporal CNN.}
\SoccerNet provides three frame feature extractors with different backbone architectures (I3D, C3D, and ResNet). 
% Each of them extracts $512$ features, which are provided with the dataset. 
% We use these pre-computed features as input of our temporal CNN. 
Each of them respectively extracts $1024$, $4096$, and $2048$ features that are further reduced to $512$ features with a Principal Component Analysis (PCA).
We use the PCA-reduced features provided with the dataset as input of our temporal CNN.
%The frame feature extractor extracts features on a per-frame basis and outputs $512$ features. These features are provided with the \SoccerNet dataset with three backbones (ResNet, C3D and I3D). The frame feature extractor is thus not trained with. These $512$ features are then used as input of our temporal CNN.
%We use the $512$ features provided for each frame with \SoccerNet as the output of the frame feature extractor and input of the temporal CNN.

The aim of the temporal CNN is to provide $Cf$ features for each frame, while mixing temporal information across the frames. It transforms an input of shape $N_F \times 512$ into an output of shape $N_F \times Cf$.

First, each frame is input to a $2$-layer MLP to reduce the dimensionality of the feature vectors of each frame. We design its architecture as: FC($128$) - ReLU - FC($32$) - ReLU. We thus obtain a set of $N_F\times 32$ features, which we note $\mathcal{F}_\text{MLP}$.

Then, $\mathcal{F}_\text{MLP}$ is input to a spatio-temporal pyramid, \ie it is input in parallel to each of the following layers of the pyramid:
\begin{itemize}
    \item Conv($8, r/7\times 32$) - ReLU
    \item Conv($16, r/3\times 32$) - ReLU
    \item Conv($32, r/2\times 32$) - ReLU
    \item Conv($64, r\times 32$) - ReLU
\end{itemize}
producing $8+16+32+64=120$ features for each frame, which are concatenated with $\mathcal{F}_\text{MLP}$ to obtain a set of $N_F\times 152$ features. 

Finally, we feed these features to a Conv($Cf, 3\times 152$) layer, which produces a set of $N_F\times Cf$ features, noted $\mathcal{F}_\text{TCNN}$.

\textbf{\color{NavyBlue}2. Segmentation module.} This module produces a segmentation score per class for each frame. It transforms $\mathcal{F}_\text{TCNN}$ into an output of dimension $N_F\times C$, through the following steps:
\begin{itemize}
    \item Reshape $\mathcal{F}_\text{TCNN}$ to have dimension $N_F\times C\times f$.
    \item Use a frame-wise Batch Normalization. 
    \item Activate with a sigmoid so that each frame has, for each class, a feature vector $v \in (0,1)^f$.
    \item For each frame, for each class, compute the distance $d$ between $v$ and the center of the unit hypercube $(0,1)^f$, \ie a vector composed of $0.5$ for its $f$ components. Hence, $d\in [0,\sqrt{f}/2]$.
    \item The segmentation score is obtained as $1-2d/\sqrt{f}$, which belongs to $[0,1]$. This way, scores close to $1$ for a class (\ie $v$ close to the center of the cube) can be interpreted as indicating that the frame is likely to belong to that class.
\end{itemize}
The segmentation scores $\zeta_\text{seg}$ output by the segmentation module thus has dimension $N_F\times C$ and is assessed through the segmentation loss $\mathcal{L}^\text{seg}$.

\textbf{\color{Green}3. Spotting module.} The spotting module takes as input $\mathcal{F}_\text{TCNN}$ and $\zeta_\text{seg}$, and outputs the spotting predictions $\hat{\textbf{Y}}$ of the network. It is composed of the following layers:
\begin{itemize}
    \item ReLU on $\mathcal{F}_\text{TCNN}$, then concatenate with $\zeta_\text{seg}$. This results in $N_F\times (Cf+C)$ features.
    \item Temporal max-pooling $3\times 1$ with a $2\times 1$ stride.
    \item Conv($32, 3\times (Cf+C)$) - ReLU
    \item Temporal max-pooling $3\times 1$ with a $2\times 1$ stride.
    \item Conv($16, 3\times 32$) - ReLU
    \item Temporal max-pooling $3\times 1$ with a $2\times 1$ stride.
    \item Flatten the resulting features, which yields $\mathcal{F}_\text{spot}$.
    \item Feed $\mathcal{F}_\text{spot}$ to a FC($2N_\text{pred}$) layer, then reshape to $N_\text{pred}\times 2$ and use sigmoid activation. This produces the confidence scores and the predicted locations for the action spots.
    \item Feed $\mathcal{F}_\text{spot}$ to a FC($CN_\text{pred}$) layer, then reshape to $N_\text{pred}\times C$ and use softmax activation on each row. This produces the per-class predictions for the action spots.
    \item Concatenate the confidence scores, predicted locations, and per-class predictions to produce the spotting predictions $\hat{\textbf{Y}}$ of shape $N_\text{pred}\times (2+C)$.
\end{itemize}
Eventually, $\hat{\textbf{Y}}$ is assessed through the action spotting loss $\mathcal{L}^\text{as}$.

%further details on TSE

%one to one matching (Figure) (YOLO-like)

\subsection{Iterative One-to-One Matching}

\begin{figure}
    \centering
    \includegraphics[width=0.95\columnwidth]{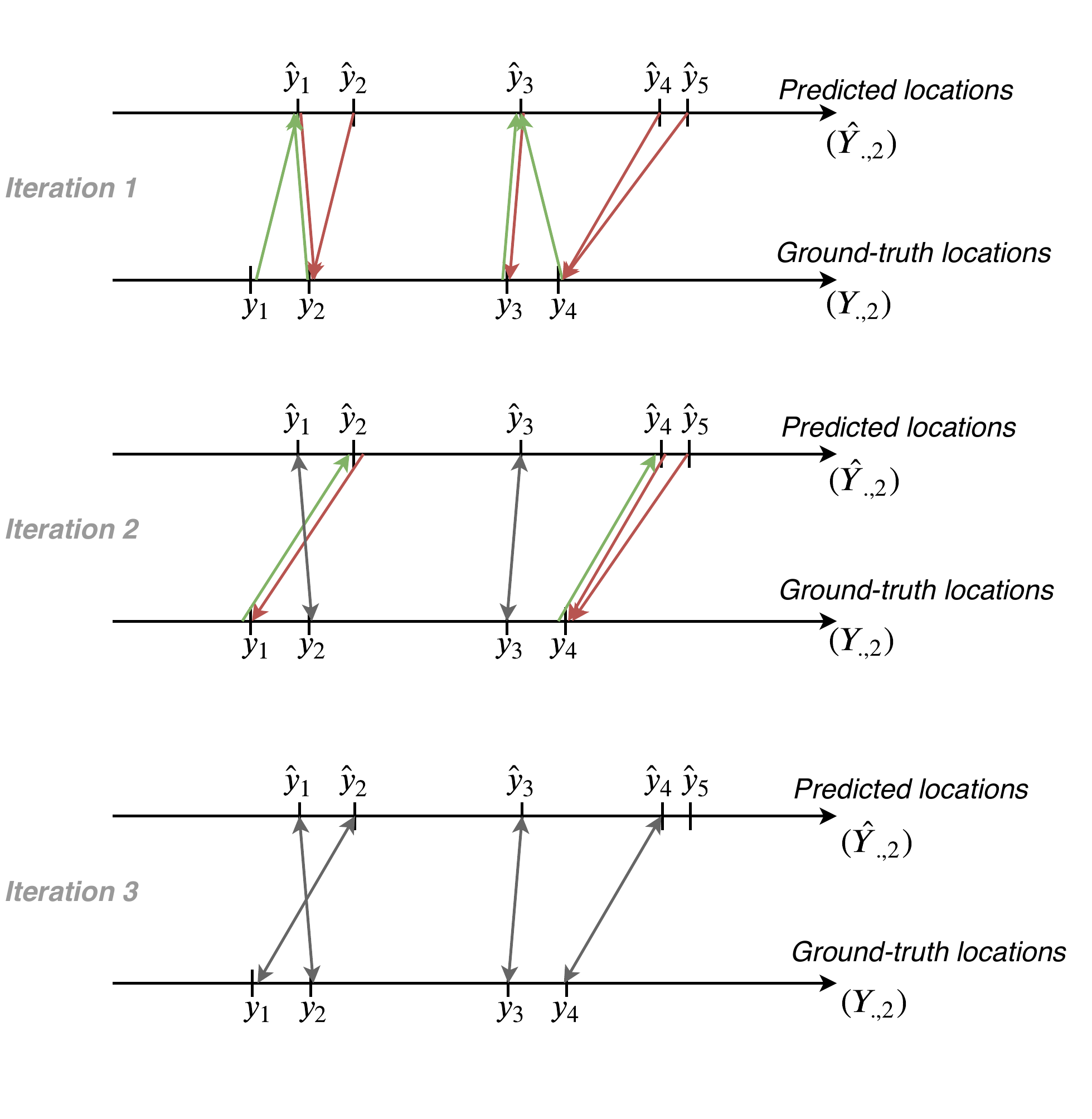}\\
    \caption{\textbf{Iterative one-to-one matching.} Example of the iterative one-to-one matching. At iteration $1$, each ground-truth location is matched with its closest predicted location (\textbf{{\color{anthogreen}green arrows}}), and vice-versa (\textbf{{\color{anthobrown}brown arrows}}). Locations that match each other are permanently matched (\textbf{\color{gray}gray arrows}), and the process is repeated with the remaining locations at iteration 2. In this case, two iterations suffice to match all the ground-truth locations with a predicted location, as evidenced by the absence of available ground-truth location for iteration 3. 
    }
    \label{fig:onetoonematching}
\end{figure}

The iterative one-to-one matching between the predicted locations $\hat{\textbf{Y}}_{\cdot,2}$ and the ground-truth locations $\textbf{Y}_{\cdot,2}$ described in the paper is illustrated in Figure~\ref{fig:onetoonematching}. It is further detailed mathematically in Algorithm~\ref{alg:matching}.%, with generic sets of real numbers $X$ and $Y$ to be matched.

\begin{algorithm}
\SetAlgoLined
\KwData{$Y, \hat{Y}$ ground-truth and predicted locations}
\KwResult{Matching couples $(y,\hat{y}) \in Y \times \hat{Y}$}
{\bf Algorithm:}\\
\While{$Y \neq \emptyset$}{
    $f: Y\to \hat{Y}: f(y) = \argmin\{|y-\hat{y}|:\ \hat{y}\in \hat{Y}\}$\;
    \For{$\hat{y} \in \hat{Y}$}{
    \If{$|f^{-1}(\{\hat{y}\})| \geq 1$}{
        $y_{\hat{y}} = \argmin\{|y-\hat{y}|:\ y\in f^{-1}(\{\hat{y}\})\}$\;
        Save matching couple $(y_{\hat{y}},\hat{y})$\;
        Remove $y_{\hat{y}}$ from $Y$ and $\hat{y}$ from $\hat{Y}$\;
        }
    }
}
\caption{Iterative matching between ground-truth and predicted locations.}
\label{alg:matching}
\end{algorithm}

%120s Chunk

\subsection{Details on the Time-Shift Encoding (TSE)}

\begin{figure}
    \centering
    \includegraphics[width=\linewidth]{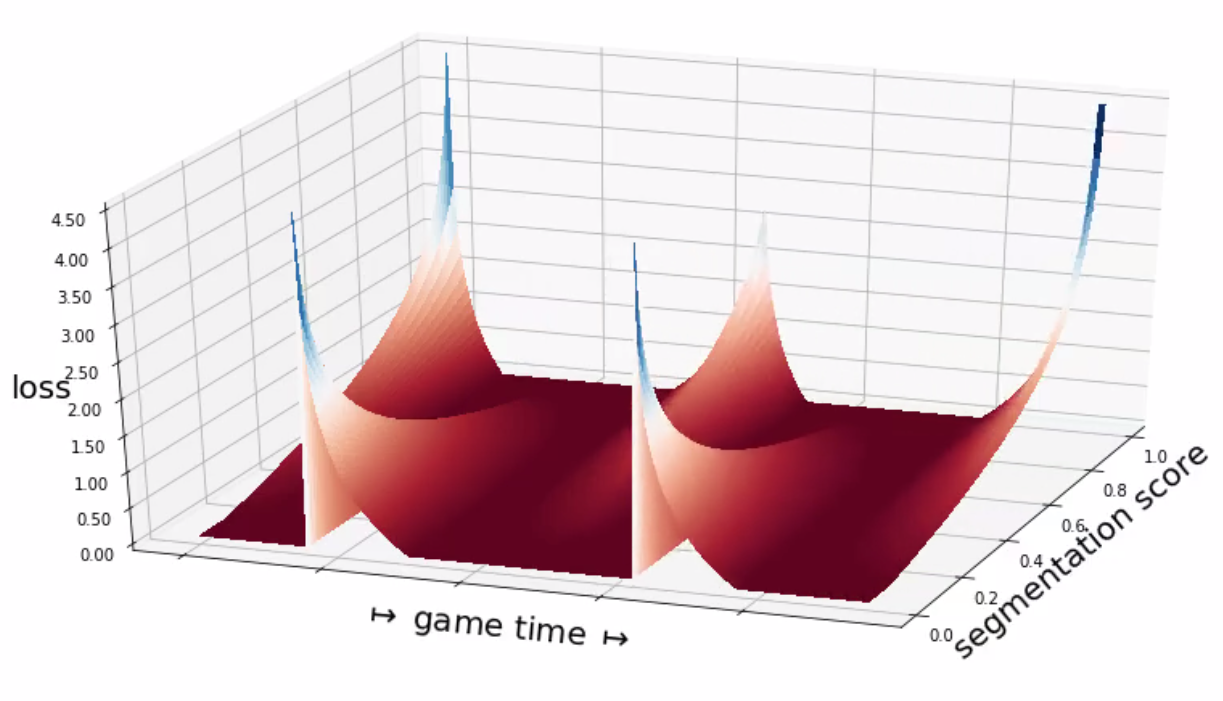}
    \caption{\textbf{Context-aware loss function (close actions).} Representation of our segmentation loss when two actions of the same class are close to each other. The loss is parameterized by the time-shift encoding of the frames and is continuous through time, except at frames annotated as actions. A video clip where we vary the location of the second action is provided with this document (\emph{3dloss.mp4}).}
    \label{fig:two-events-loss}
\end{figure}

The time-shift encoding (TSE) described in the paper is further detailed below. We note $s^c(x)$ the TSE of frame $x$ related to class $c$.

We denote $s^c_p$ (resp. $s^c_f$) the difference between the frame index of $x$ and the frame index of its closest past (resp. future) ground-truth action of class $c$. They constitute the time-shifts of $x$ from its closest past and future ground-truth actions of class $c$, expressed in number of frames (\ie if frames $9$ and $42$ are actions of class $c$, then frame $29$ has $s^c_p=29-9=20$ and $s^c_f=29-42=-13$). %Hence, $s^c_p$ (resp. $s^c_f$) is computed as the difference between the frame index of $x$ and the frame index of its closest past (resp. future) ground truth action of class $c$. 
We set $s^c_p=0$ for a frame corresponding to a ground-truth action of class $c$, thus ensuring the relations $s^c_f < 0 \leq s^c_p$. 
The TSE $s^c(x)$ is defined as the time-shift among $\{s^c_p, s^c_f\}$ related to the action that has the dominant influence on $x$. %We set $s^c(x)=s^c_p$ if one of the following conditions based on our context slicing is met: 
%\textbf{1.} $x$ is located just after the past action; 
%\textbf{2.} $x$ is located in the transition zone after the past action, but located far before the future action; 
%\textbf{3.} $x$ is located in the transition zone after the past action, and located in the transition zone before the future action, but $x$ is closer to the past action than the future action. %well, technically, not really, there is that proportion story  $\frac{s_p-K^c_3}{K^c_4-K^c_3} < \frac{K^c_2-s_f}{K^c_2-K^c_1}$, but I guess we can keep this for supplementary material...
%In all other cases, the future action influences more $x$ than the past action, hence $s^c(x)=s^c_f$.
The rules used to determine which time-shift is selected are the following:
\begin{itemize}
    \item if $s^c_p < K^c_3$: keep $s^c_p$, because $x$ is located \emph{just after} the past action, which still strongly influences $x$. %(\eg replays, celebrations).
    \item if $K^c_3 \leq s^c_p < K^c_4$: $x$ is in the \emph{transition zone} after the past action, whose influence weakens, thus the decision depends on how far away is the future action:
    \begin{itemize}
        \item if $s^c_f \leq K^c_1$: keep $s^c_p$, because $x$ is located \emph{far before} the future action, which does not yet influence $x$.
        \item if $s^c_f > K^c_1$:
        The future action may be close enough to influence $x$:
        \begin{itemize}
            \item if $\frac{s^c_p-K^c_3}{K^c_4-K^c_3} < \frac{K^c_2-s^c_f}{K^c_2-K^c_1}$: keep $s^c_p$, %because $x$ is closer to the JA region of the past action than it is to the JB region of the future action, proportionally to the size of the regions. 
            because $x$ is closer to the \emph{just after} region of the past action than it is to the \emph{just before} region of the future action, with respect to the size of the transition zones. %hence the influence of the past action is still larger than the influence of the future action.%, even though both of them may impact the frame.
            \item else: keep $s^c_f$, because the future action influences $x$ more than the past action.
        \end{itemize}
    \end{itemize}
    \item if $s^c_p \geq K^c_4$: keep $s^c_f$, because $x$ is located \emph{far after} the past action, which does not influence $x$ anymore.
\end{itemize}
For completeness, let us recall the following details mentioned in the main paper.
If $x$ is both located \emph{far after} the past action and \emph{far before} the future action, selecting either of the two time-shifts has the same effect in our loss. Furthermore, for the frames located either before the first or after the last annotated action of class $c$, only one time-shift can be computed and is thus set as $s^c(x)$. Finally, if no action of class $c$ is present in the video, then we set $s^c(x)=K^c_1$ for all the frames. This induces the same behavior in our loss as if they were all located far before their closest future action.
%The result gives the time-shift encoding $s^c(x)$ of $x$ related to class $c$. %Repeating the process for all the classes gives the time-shift encoding $S$ of the frame $x$. %This process is repeated for all the videos, classes, and frames. 

The TSE is used to shape our novel context-aware loss function for the temporal segmentation module. The cases described above ensure the temporal continuity of the loss, regardless of the proximity between two actions of the same class, excepted at frames annotated as ground-truth actions. This temporal continuity can be visualized in Figure~\ref{fig:two-events-loss}, which shows a representation of $\tilde{L}(p,s)$ (analogous to Figure~\ref{fig:Loss}) when two actions are close to each other. It is further illustrated in the video clip \emph{3dloss.mp4} provided with this document, where we gradually vary the location of the second action. For each location of the second action, the TSE of all the frames is re-computed, and so is the loss.

\subsection{Extra Analyses}

\begin{figure}
    \centering
    \includegraphics[width=0.95\columnwidth]{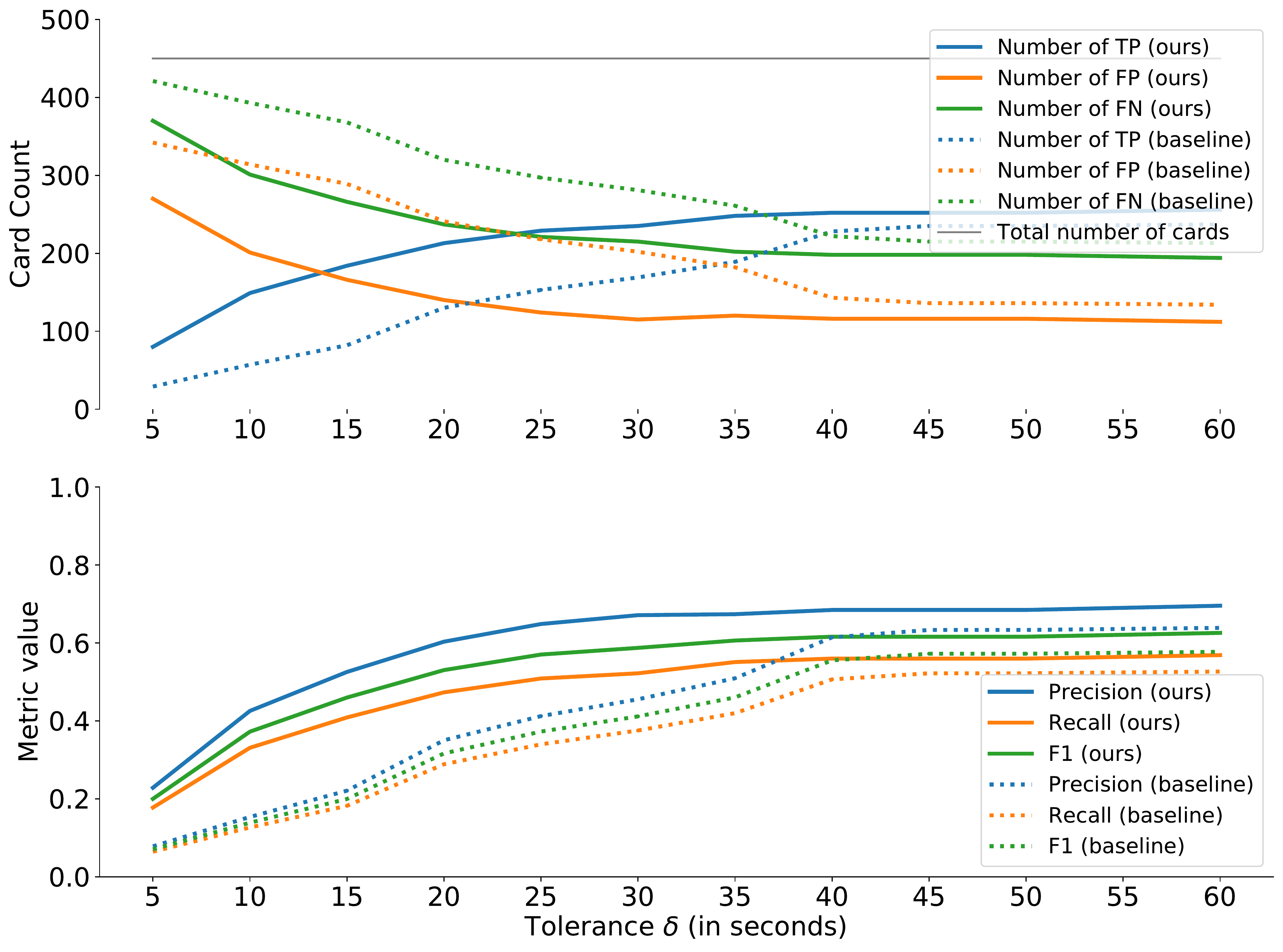}\\
    \caption{\textbf{Per-class results (cards).} A prediction of class \emph{card} is a {\color{anthoblue}\textbf{true positive (TP)}} with tolerance $\delta$ when it is located at most $\delta/2$ seconds from a ground-truth card. The baseline results are obtained from the best model of~\cite{Giancola_2018_CVPR_Workshops}. Our model spots most cards within $15$ seconds around the ground truth ($\delta=30$ seconds). 
    }
    \label{fig:cardmetrics}
\end{figure}

\begin{figure}
    \centering
    \includegraphics[width=0.95\columnwidth]{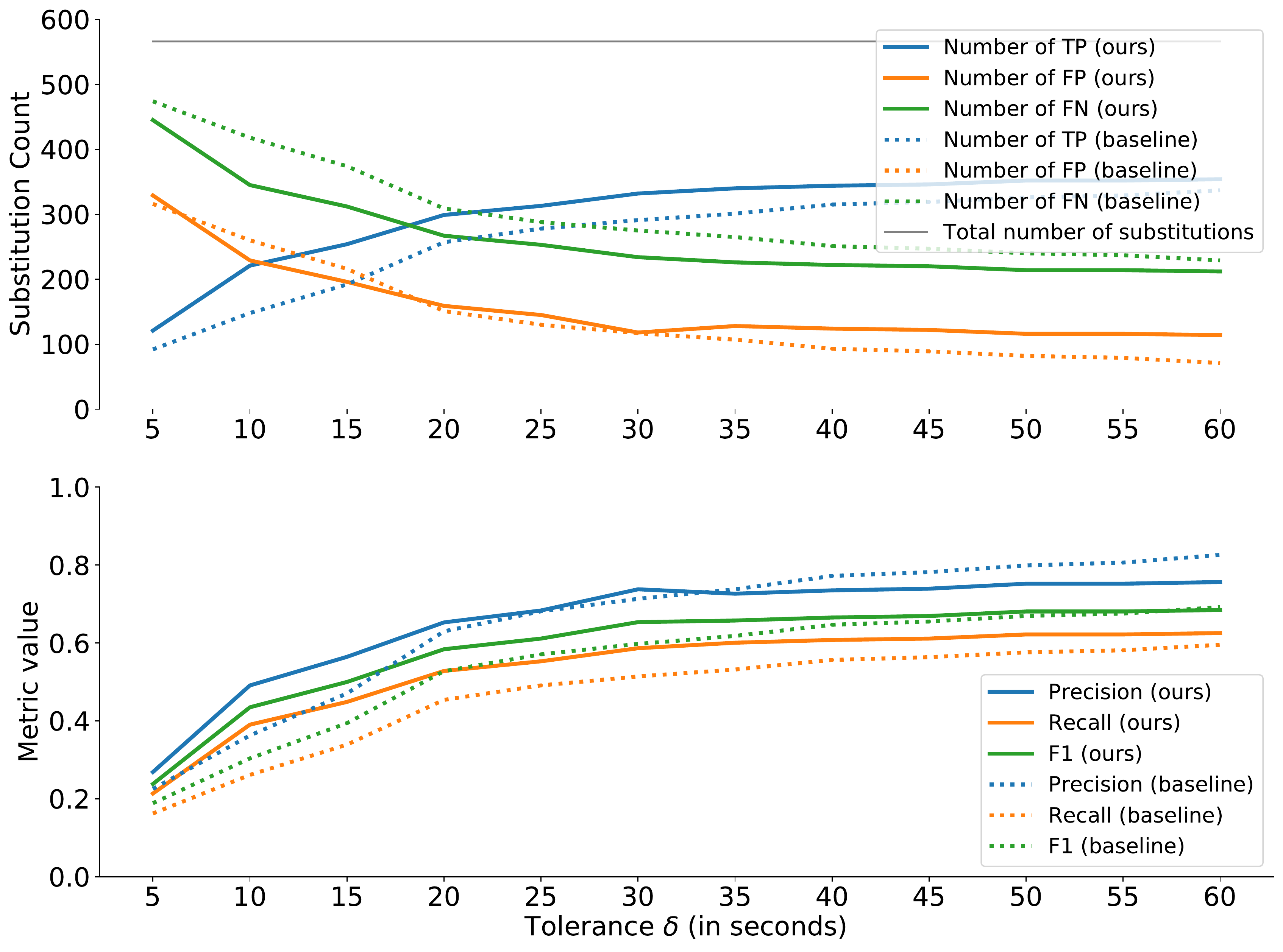}\\
    \caption{\textbf{Per-class results (substitutions).} A prediction of class \emph{substitution} is a {\color{anthoblue}\textbf{true positive (TP)}} with tolerance $\delta$ when it is located at most $\delta/2$ seconds from a ground-truth substitution. The baseline results are obtained from the best model of~\cite{Giancola_2018_CVPR_Workshops}. Our model spots most substitutions within $15$ seconds around the ground truth ($\delta=30$ seconds). 
    }
    \label{fig:substitutionmetrics}
\end{figure}

\mysection{Per-class results.} As for the class \emph{goal} in Figure~\ref{fig:goalmetrics} of the main paper, Figures~\ref{fig:cardmetrics} and~\ref{fig:substitutionmetrics} display the number of TP, FP, FN and the precision, recall and $F_1$ metrics for the classes \emph{card} and \emph{substitution} as a function of the tolerance $\delta$ allowed for the localization of the spots. 

Figure~\ref{fig:cardmetrics} shows that most cards can be efficiently spotted by our model within $15$ seconds around the ground truth ($\delta=30$ seconds). We achieve a precision of $66\%$ for that tolerance. The previous baseline plateaus within $20$ seconds ($\delta=40$ seconds) and still has a lower performance. %Goals appear to be more easily spotted, which may be due to

Figure~\ref{fig:substitutionmetrics} shows that most substitutions can be efficiently spotted by our model within $15$ seconds around the ground truth ($\delta=30$ seconds). We achieve a precision of $73\%$ for that tolerance. The previous baseline reaches a similar performance for that tolerance, and reaches $82\%$ within $60$ seconds ($\delta=120$ seconds) around the ground truth. 

Except for the precision metric for the substitutions with tolerances larger than $20$ seconds, our model outperforms the previous baseline of \SoccerNet. As mentioned in the paper, for goals, many visual cues facilitate their spotting, \eg multiple replays, particular camera views, or celebrations from the players and from the public. Cards and substitutions are more difficult to spot since the moment the referee shows a player a card and the moment a new player enters the field to replace another are rarely replayed (\eg for cards, the foul is replayed, not the sanction). Also, the number of visual cues that allow their identification is reduced, as these actions generally do not lead to celebrations from the players or the public. Besides, cards and substitutions may not be broadcast in full screen, as they are sometimes merely shown from the main camera and are thus barely visible. Finally, substitutions occurring during the half-time are practically impossible to spot, as said in the main paper.


\begin{figure}
    \centering
    \includegraphics[width=\linewidth]{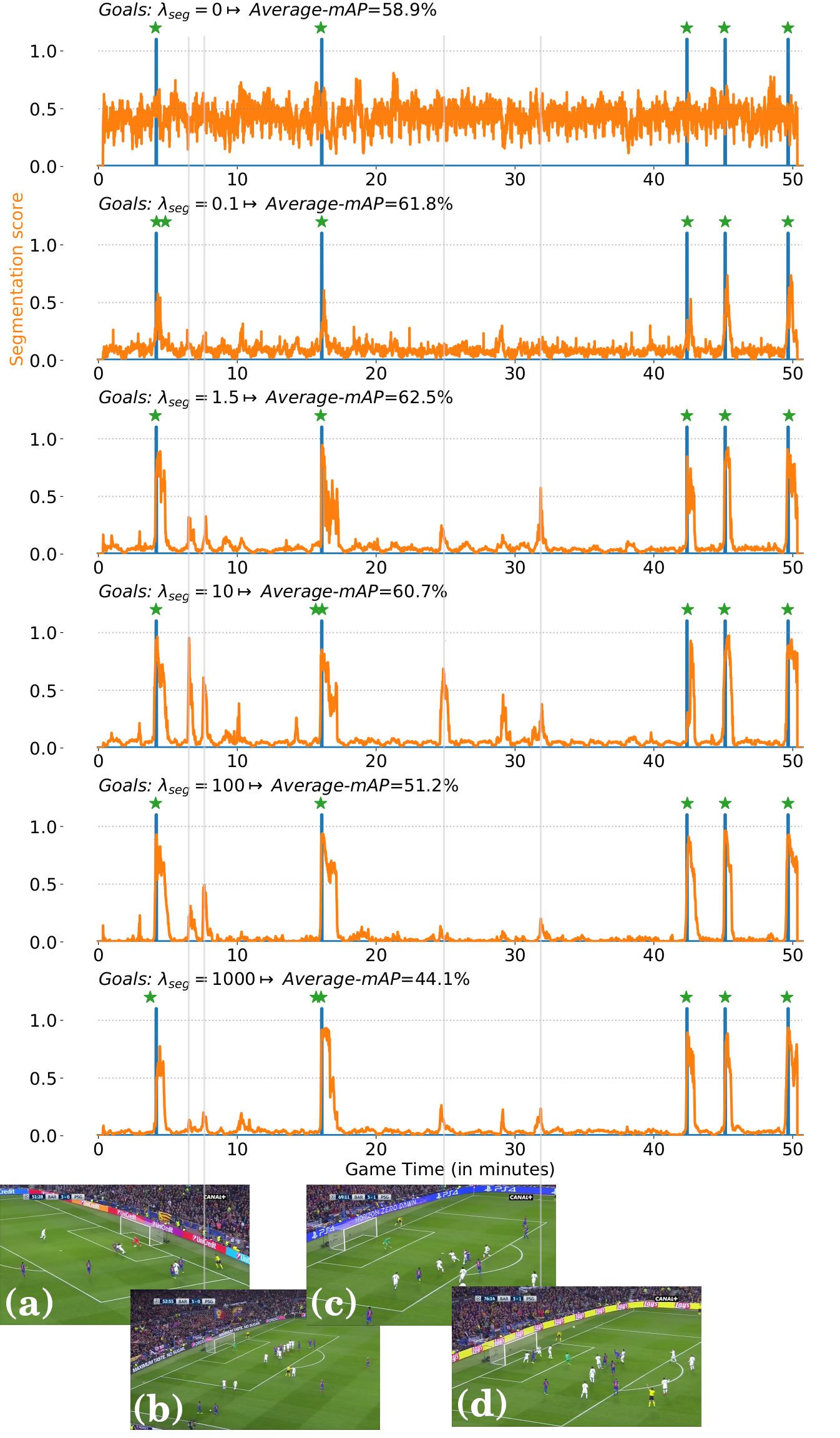}
    \caption{\textbf{Influence of $\lambda^\text{seg}$} on the segmentation and spotting results of the second half of the famous ``Remuntada" match, Barcelona - PSG, for the class \emph{goal}, for different values of $\lambda^\text{seg}$. The best Average-mAP for the spotting task is located around $\lambda^\text{seg}=1.5$, while the best value for spotting unannotated goal opportunities might be around $\lambda^\text{seg}=10$. For this value, several meaningful \emph{goal opportunities} have a high \textbf{\color{orange}segmentation score}: \textbf{(a)} a shot on a goal post, \textbf{(b)} a free kick, \textbf{(c)} lots of dribbles in the rectangle, and \textbf{(d)} a headshot right above the goal.
    }
    \label{fig:lambdaAnalysis}
\end{figure}

\mysection{Segmentation loss analysis.} We provide a supplementary analysis on the $\lambda^\text{seg}$ parameter, which balances the segmentation loss and the action spotting loss in Equation~\ref{eq:LossFinal} of the main paper.
We fix different values of $\lambda^\text{seg}$ and train a network for each value. We show the segmentation scores on one game for the \emph{goal} class in Figure~\ref{fig:lambdaAnalysis}. We also display the Average-mAP for the whole test set for the different values of $\lambda^\text{seg}$.

It appears that extreme values of $\lambda^\text{seg}$ substantially influence both the action spotting performance and the segmentation curves, hence the automatic highlights generation. Small values (\ie $\lambda^\text{seg}\leq 0.1$) produce a useless segmentation for spotting the interesting unannotated \emph{goal opportunities}. This is because the loss does not provide a sufficiently strong feedback for the segmentation task as it does not penalize enough the segmentation scores. These values of $\lambda^\text{seg}$ also lead to a decrease in the Average-mAP for the action spotting task, as already observed in the ablation study presented in the main paper. Moreover, very large values ($\lambda^\text{seg}\geq 100$) penalize too much the unannotated goal opportunities, for which the network is then forced to output very small segmentation scores. Such actions are thus more difficult to retrieve for the production of highlights. These values of $\lambda^\text{seg}$ also lead to a large decrease in the Average-mAP for the action spotting task, as the feedback of the segmentation loss overshadows the feedback of the spotting loss. Finally, it seems that for $\lambda^\text{seg}\in[1,10]$, the spotting performance is high while providing informative segmentation scores on \emph{goal opportunities}. These values lead to the spotting of several \emph{goal opportunities}, shown in Figure~\ref{fig:lambdaAnalysis}, which might be included in the highlights automatically generated for this match by the method described in the main paper.

%\mysection{DETAD Analysis.}

%We ran the DETAD diagnosis tool on the ActivityNet experiments, to highlight the origin of the improvement made.
%We show here that we commit less error in Background, Confusion, Wrong Label and Double Detection.

%\begin{figure}
%    \centering
%    \includegraphics[width=0.485\linewidth]{imgs/DETAD_Baseline_20.pdf}
%    \includegraphics[width=0.485\linewidth]{imgs/DETAD_NoCaps_20.pdf}
%    \includegraphics[width=0.485\linewidth]{imgs/DETAD_Baseline_30.pdf}
%    \includegraphics[width=0.485\linewidth]{imgs/DETAD_NoCaps_30.pdf}
%    \caption{
%    \textbf{DETAD False positive analysis}
%    for the Baseline BMN (left), and
%    for the Baseline BMN with our loss (right)
%    }
%    \label{fig:DETAD}
%\end{figure}

\mysection{Comments on improvements on ActivityNet.}
In Table~3, we report the averages over samples of $20$ results for each metric, that we further analyze statistically below for the Av.-mAP.
First, following D'Agostino's normality test, we can reasonably assume that the samples are normally distributed, since we obtain $p$-values $>0.1$ ($0.28$ for BMN and $0.24$ for ours respectively). The standard deviations of the samples are $0.08\%$ and $0.07\%$. Since the difference between the averages is $0.38\%$, the normal distributions overlap beyond two standard deviations from their centers, which shows that our improvements are beyond noise domain. Furthermore, Bartlett's test for equal variances gives a $p$-value of $0.62$ ($> 0.1$), which allows us to use Student's $t$-test to check whether the two samples can be assumed to have the same mean or not. We obtain a $p$-value of \num{2.3e-18}, which strongly indicates that our results are significantly different from those of BMN and hence confirm the significant improvement.
For the AR@100 and AUC, similar analyses give final $p$-values of \num{7.4e-3} and \num{9.8e-2}, which corroborates the statistical significance of our improvements.

\subsection{Extra Actions and Highlights Generation}

\begin{figure}
    \centering
    \includegraphics[width=\linewidth]{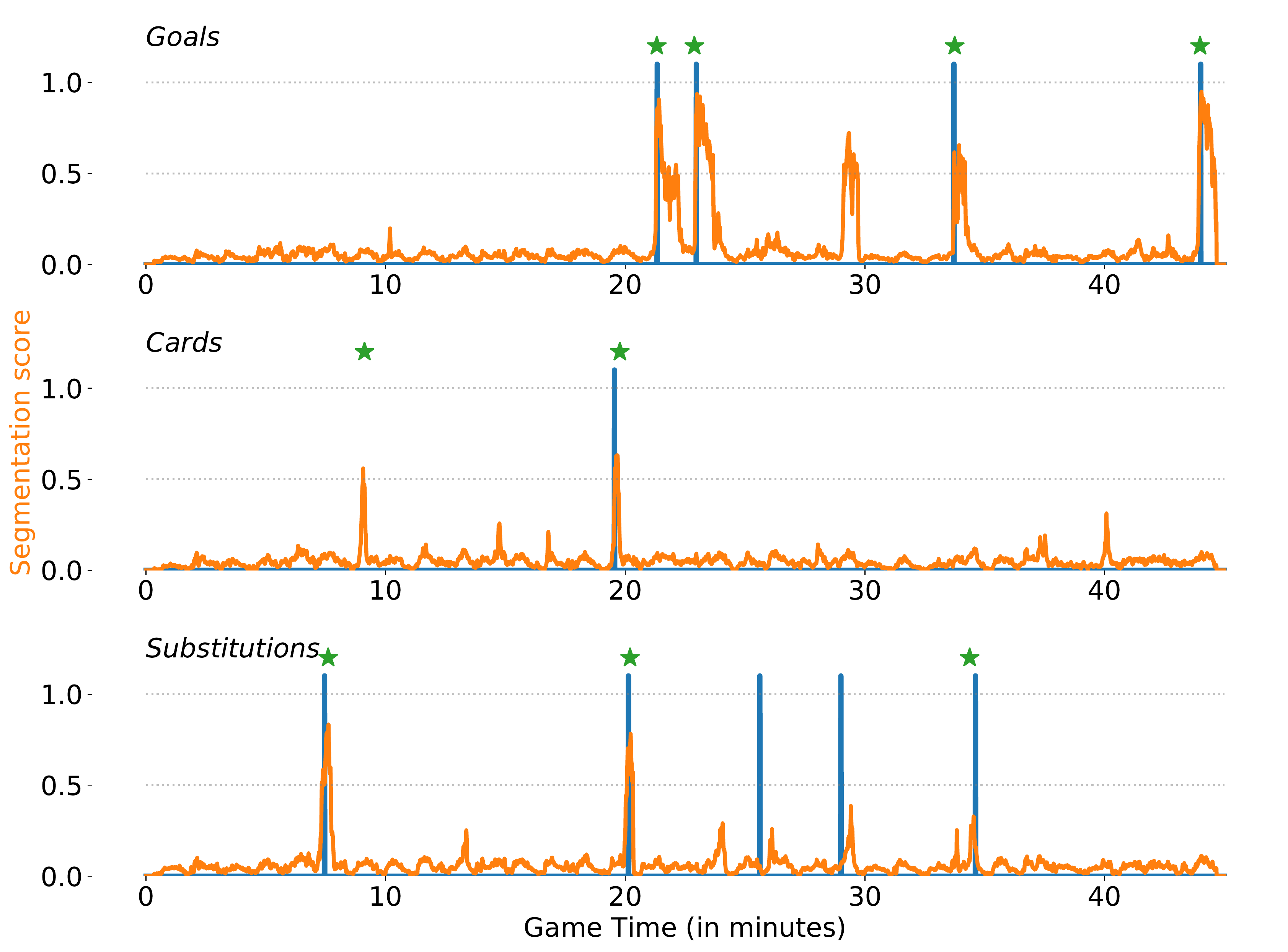}
    \caption{
    \textbf{Extra action spotting and segmentation results.} These results are obtained on the second half of the match Barcelona - Espanyol in December 2016. {\color{anthoblue} \textbf{Ground truth actions}}, {\color{anthoorange}\textbf{temporal segmentation curves}}, and {\color{anthogreen}\textbf{spotting results (green stars)}} are illustrated. Unannotated actions can be identified and included in the highlights using our segmentation. For example, a goal opportunity occurs around the $29^{\text{th}}$ minute. A false positive spot for a card is predicted by our network around the $9^{\text{th}}$ minute. As it corresponds to a severe unsanctioned foul, it is fine for our automatic highlights generator to include it in the summary of the match.
    }
    \label{fig:match_189}
\end{figure}

Figure~\ref{fig:match_189} shows additional action spotting and segmentation results. We can identify actions that are unannotated but display high segmentation scores such as goal opportunities and unsanctioned fouls. A goal opportunity around the $29^{\text{th}}$ minute can be identified through the segmentation results. Besides, a false positive spot (green star) for a card is predicted around the $9^{\text{th}}$ minute, further supported by a high segmentation score. A manual inspection reveals that a severe unsanctioned foul occurs at this moment. The automatic highlights generator presented in the main paper would include it in the summary of the match. Even though this foul does not lead to a card for the offender, the content of this sequence corresponds to an interesting action that would be tolerable in a highlights video.

%As it corresponds to an unsanctioned foul, it is fine for our automatic highlights generator to include it in the summary of the match.  %The highlights videos automatically generated for these two matches are provided with this document.

%\begin{figure}
%    \centering
%    \includegraphics[width=\linewidth]{imgs/segmentation_detection103.pdf}
%    \caption{
%    \textbf{Action spotting and segmentation}  for the second half of the match Dortmund - Bayer Leverkusen in March 2017. {\color{anthoblue} \textbf{Ground truth actions}}, {\color{anthoorange}\textbf{temporal segmentation curves}}, and {\color{anthogreen}\textbf{spotting results (green stars)}} are illustrated. Unannotated actions can be identified and included in the highlights using our segmentation, such as a goal opportunity around the $34^{\text{th}}$ minute.
%    }
%    \label{fig:match_103}
%\end{figure}

\begin{figure}
    \centering
    \includegraphics[width=\linewidth]{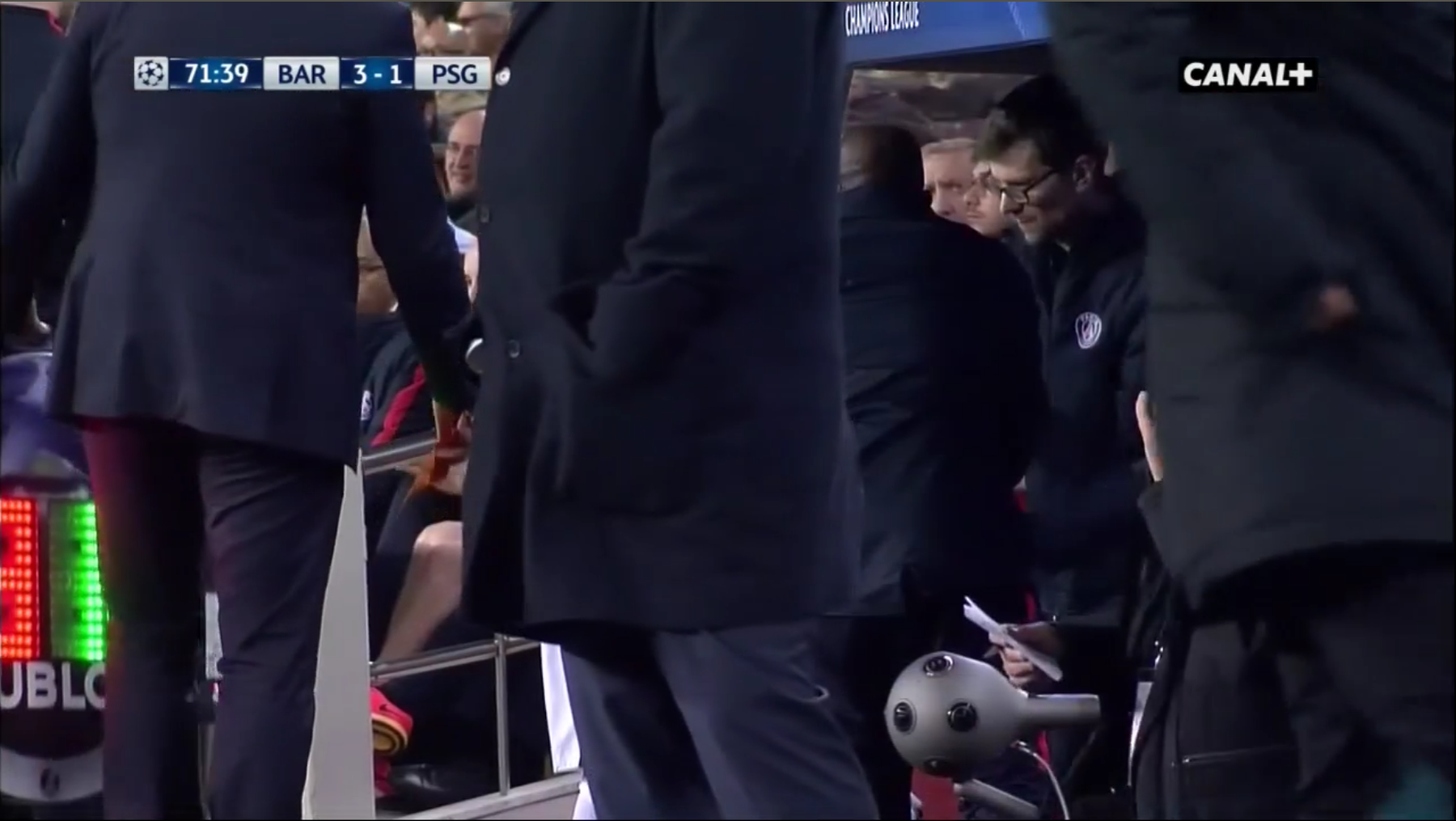}
    \caption{
    \textbf{False positive spot of a substitution} for the second half of the famous ``Remuntada" match, Barcelona - PSG, in March 2017. The LED panel used to announce substitutions is visible on the left, which presumably explains why the network predicted the sequence around this frame as a substitution.
    }
    \label{fig:FP_substitution}
\end{figure}

Figure~\ref{fig:FP_substitution} shows a frame for which our network provides a high segmentation score and a false positive spot around the $26^\text{th}$ minute (\ie $71^\text{st}$ minute of the match) for \emph{substitutions} in Figure~\ref{fig:predsandsegs} of the main paper. We can see that the LED panel used by the referee to announce substitutions is visible on the frame. This may indicate that the network learns, quite rightly, to associate this panel with substitutions. As a matter of fact, at this moment, even the commentator announces that a substitution is probably imminent.

% ADD IT ONLY IF NECESSARY
% \subsection{DETAD Analysis}

% We run DETAD on ActivityNet to diagnosis the errors \cite{alwassel2018diagnosing}.

% \begin{figure}[h]
%     \centering
%     \includegraphics[width=\linewidth]{results/detad/anet_winner/false_positive_analysis.pdf}
%     \includegraphics[width=\linewidth]{results/detad/baseline/1/false_positive_analysis.pdf}
%     \includegraphics[width=\linewidth]{results/detad/nocaps/8/false_positive_analysis.pdf}
%     \caption{DETAD Results on ActivityNet v1.3: False Positive Analysis for ANET18 Winner, BMN, BMN+Ours}
%     \label{fig:DETADFalsePositiveAnalysis}
% \end{figure}

% \begin{figure}[h]
%     \centering
%     \includegraphics[width=\linewidth]{results/detad/anet_winner/false_negative_analysis.pdf}
%     \includegraphics[width=\linewidth]{results/detad/baseline/1/false_negative_analysis.pdf}
%     \includegraphics[width=\linewidth]{results/detad/nocaps/8/false_negative_analysis.pdf}
%     \caption{DETAD Results on ActivityNet v1.3: False Negative Analysis for ANET18 Winner, BMN, BMN+Ours}
%     \label{fig:DETADFalseNegativeAnalysis}
% \end{figure}

% \begin{figure}[h]
%     \centering
%     \includegraphics[width=\linewidth]{results/detad/anet_winner/sensitivity_analysis.pdf}
%     \includegraphics[width=\linewidth]{results/detad/baseline/1/sensitivity_analysis.pdf}
%     \includegraphics[width=\linewidth]{results/detad/nocaps/8/sensitivity_analysis.pdf}
%     \caption{DETAD Results on ActivityNet v1.3: Sensitivity Analysis for ANET18 Winner, BMN, BMN+Ours}
%     \label{fig:DETADSensitivityAnalysis}
% \end{figure}

\end{document}